\documentclass[10pt,journal,compsoc]{IEEEtran}
\ifCLASSOPTIONcompsoc
  \usepackage[nocompress]{cite}
\else
  \usepackage{cite}
\fi
\ifCLASSINFOpdf
\else
\fi
\usepackage{epsfig}
\usepackage{graphicx}
\usepackage{amsmath}
\usepackage{amssymb}

\usepackage{url}
\usepackage{comment}
\usepackage{color}
\usepackage{algorithm}
\usepackage{algpseudocode}
\usepackage{subfigure}
\usepackage{enumitem}
\usepackage{grffile}

\usepackage{makecell}
\usepackage{graphbox}
\usepackage{booktabs}
\usepackage{bm}
\usepackage{multirow}
\usepackage{wrapfig}
\usepackage{xcolor}
\usepackage{color, colortbl}
\definecolor{Gray}{gray}{0.9}
\definecolor{White}{rgb}{1,1,1}
\usepackage[english]{babel}
\begin{document}
\title{General Adversarial Defense Against Black-box Attacks via Pixel Level and Feature Level Distribution Alignments}

\author{Xiaogang Xu, Hengshuang Zhao, Philip Torr, and Jiaya Jia,~\IEEEmembership{Fellow,~IEEE}% <-this % stops a space
	\IEEEcompsocitemizethanks{\IEEEcompsocthanksitem X.~Xu and J.~Jia are with the Department of Computer Science and Engineering, The Chinese University of Hong Kong. %\protect\\
		%E-mail: xgxu@cse.cuhk.edu.hk, leojia@cse.cuhk.edu.hk
		E-mail: \{xgxu, leojia\}@cse.cuhk.edu.hk.
		%\IEEEcompsocthanksitem Hengshuang Zhao is with The University of Hong Kong.
		\IEEEcompsocthanksitem H.~Zhao is with the Department of Computer Science, The University of Hong Kong.
		E-mail: hszhao@cs.hku.hk.
		\IEEEcompsocthanksitem P.~Torr is with the Department of Engineering  Science, University of Oxford.
		E-mail: philip.torr@eng.ox.ac.uk.
}}


\markboth{Technical Report}
{Shell \MakeLowercase{\textit{et al.}}: Bare Demo of IEEEtran.cls for Computer Society Journals}



\IEEEtitleabstractindextext{%
\begin{abstract}
Deep Neural Networks (DNNs) are vulnerable to the black-box adversarial attack that is highly transferable.
This threat comes from the distribution gap between adversarial and clean samples in feature space of the target DNNs. In this paper, we use Deep Generative Networks (DGNs) with a novel training mechanism to eliminate the distribution gap.
The trained DGNs align the distribution of adversarial samples with clean ones for the target DNNs by translating pixel values. Different from previous work, we propose a more effective pixel-level training constraint to make this achievable, thus enhancing robustness on adversarial samples. Further, a class-aware feature-level constraint is formulated for integrated distribution alignment. Our approach is general and applicable to multiple tasks, including image classification, semantic segmentation, and object detection. We conduct extensive experiments on different datasets. Our strategy demonstrates its unique effectiveness and generality against black-box attacks.
\end{abstract}

}

\maketitle

\IEEEdisplaynontitleabstractindextext

% \fi
\IEEEpeerreviewmaketitle

\IEEEraisesectionheading{\section{Introduction}\label{sec:introduction}}

\IEEEPARstart{M}{ost} deep learning models are vulnerable to adversarial samples \cite{szegedy2013intriguing,goodfellow2014explaining,arnab2018robustness,xie2017adversarial} that are maliciously generated to fool the target model by adding adversarial perturbation to the original input. Such perturbations are imperceptible to the human visual system, severely threatening real-world deep learning applications of face recognition \cite{dong2019efficient,joshi2019semantic}, self-driving
cars \cite{jia2019fooling,kong2020physgan}, etc.
It is hard to have full knowledge of the target model in practice, and attack often adopts black-box mechanism, utilizing the transferability of adversarial samples.
In this paper, we focus on defending such attack, following the setting of \cite{naseer2020self,theagarajan2020defending}.

To safeguard DNNs from black-box adversarial attack, a major class of adversarial defense applies input transformation to the input samples before they are processed by target DNNs.
A fundamental strategy is to use image processing operations without learning (e.g., image compression \cite{liu2019feature} and quilting \cite{guo2017countering}).
Alternatively, learning-based methods \cite{mustafa2019image,liao2018defense,naseer2020self} train Deep Generative Networks (DGNs) to accomplish such transformation, resulting in higher robustness. The network is trained with adversarial/clean samples as input, and synthesizes output, on which the target model works well.
The essence of these methods is to weaken difference between adversarial and the corresponding clean samples via updating their pixel values with the network.
Current approaches to train such DGNs can be divided into three categories: 1) setting pixel-level constraints for reduction of the distance between adversarial and clean samples \cite{shen2017ape,mustafa2019image}; 2) adopting constraints in the feature level of target models \cite{liao2018defense}; 3) and employing constraints in the pixel and feature level \cite{naseer2020self,li2020enhancing}.
All these methods ignore the overall distribution alignment in feature spaces of target models, which is a potential problem affecting robustness.

\begin{figure}[t]
	\centering
	\resizebox{1.0\linewidth}{!}{
		\begin{tabular}{cccc}
			\includegraphics[width=0.25\linewidth]{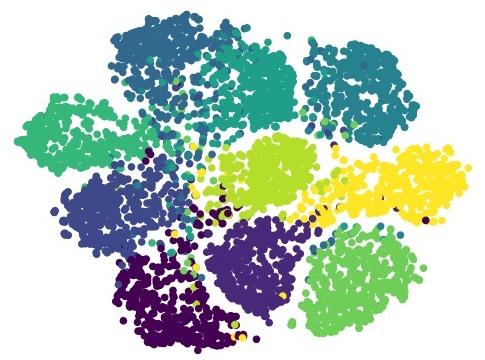}&
			\includegraphics[width=0.25\linewidth]{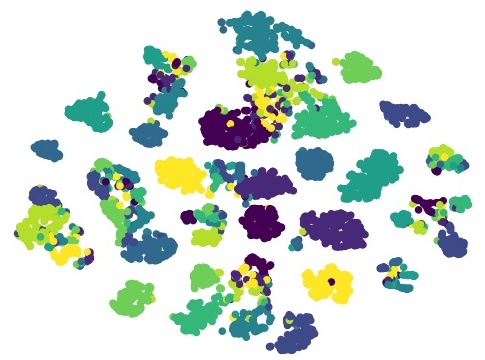}&
			\includegraphics[width=0.25\linewidth]{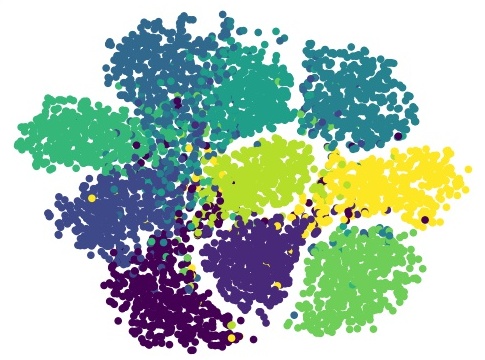}&
			\includegraphics[width=0.25\linewidth]{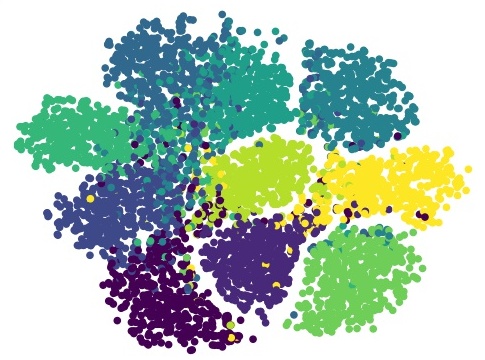}\\
			(a) $x^c$ in $\mathcal{O}$ & (b) $x^a$ in $\mathcal{O}$& (c) $\widehat{x}^c$ in $\mathcal{O}$   &(d) $\widehat{x}^a$ in $\mathcal{O}$
	\end{tabular}}
	\vspace{-0.1in}
	\caption{t-SNE visualizations in the feature space of the target model on CIFAR10  \cite{krizhevsky2009learning}. The target model $\mathcal{O}$ has proper distribution for clean samples $x^c$ and disordered distribution for adversarial samples $x^a$. Our trained generator $\mathcal{G}$ turns $x^c$ into $\widehat{x}^c$, $x^a$ into $\widehat{x}^a$ for correct distribution.}
	\vspace{-0.15in}
	\label{fig:short}
\end{figure}

\begin{figure*}[t]
	\begin{center} 
		\includegraphics[width=1.0\linewidth]{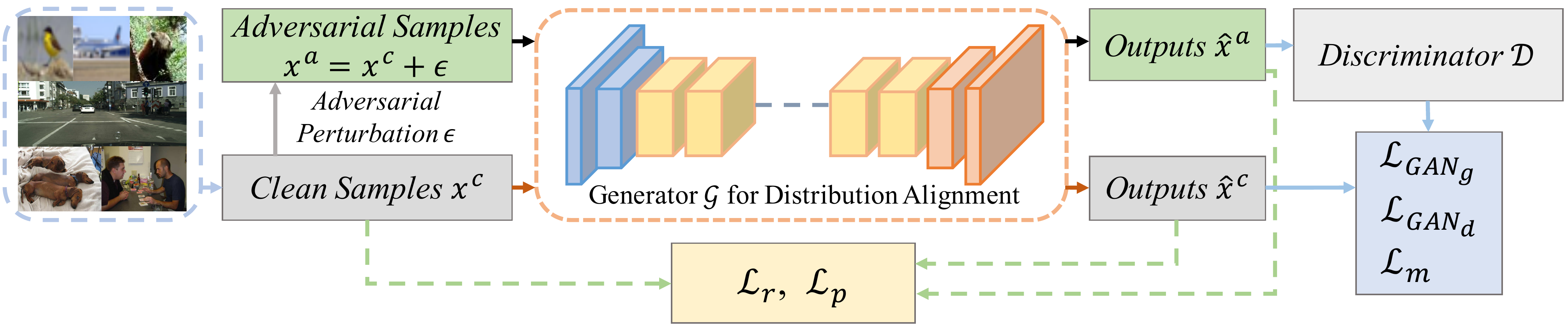}\\
		(a) Pixel level training constraints\\
		\includegraphics[width=1.0\linewidth]{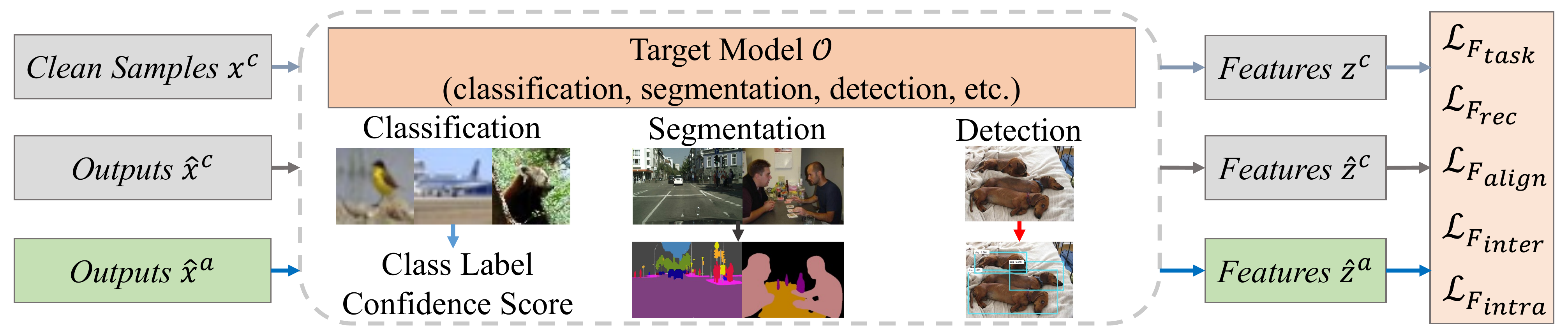}\\
		(b) Feature level training constraints\\
	\end{center}
	\vspace{-0.1in}
	\caption{Our overall framework for the training of deep generative network $\mathcal{G}$. To align the distribution of clean samples $x^c$ and adversarial samples $x^a$ for the target model, the training constraints are set in the pixel and feature levels of the target model $\mathcal{O}$.}
	\vspace{-0.1in}
	\label{fig:framework}
\end{figure*}

In this paper, we train DGNs to protect target models via aligning the distribution of clean and adversarial samples in the feature spaces of the target models.
Compared with existing methods, novel training constraints are introduced in the pixel and feature levels. In the pixel level, we match adversarial samples with clean ones in the output space of DGNs. While in the feature level, we design a class-aware constraint by aligning the central feature of clean and adversarial samples within each class, maximizing the inter-class distance as well as minimizing the intra-class distance for all categories.
Especially, our trained DGNs can be generalized to the protection of models that have not appeared during training.

DGNs trained with our method align the distribution of adversarial samples to clean ones, as exhibited in Fig.~\ref{fig:short}.
By design, our defense is general for several high-level computer vision tasks. We apply it to image classification, semantic segmentation, and object detection, with the help of diverse datasets, models, and attack. 

In summary, our contribution is the following.
\begin{itemize}
	\item Novel input transformation strategy to achieve defense against black-box attacks by distribution alignment for clean and adversarial samples, blocking the transferability of unseen adversarial samples effectively.
	\item New training constraints in both pixel and feature levels of target models. 
	\item Extensive experiments on various tasks. Our method yields high robustness, effectiveness, and generality. 
\end{itemize}

\section{Related Work}
\label{relate}

\noindent\textbf{Adversarial attack.} 
Adversarial attack involves white-box attack \cite{athalye2018robustness,goodfellow2014explaining}, where attackers have full knowledge of the target model and the defense strategy; gray-box attack \cite{guo2017countering}, where attackers have access to the target model while no access to the defense strategy; black-box attack \cite{papernot2017practical}, where attackers do not know the target model and the defense, and often use adversarial samples' transferability for attack.
Existing attacks carried out on the classification task usually compute or simulate the gradient information of target models \cite{goodfellow2014explaining,tramer2017ensemble,dong2018boosting,kurakin2016adversarial}. Meanwhile, semantic segmentation \cite{xie2017adversarial,metzen2017universal,arnab2018robustness} and object detection networks \cite{xie2017adversarial,li2018robust,song2018physical,lu2017adversarial,wei2018transferable,liu2018dpatch} are also vulnerable to adversarial attack. 
It is common sense that the black-box attack is more common than the white-box and gray-box attack for real-world applications, and it is worth exploring how to defend such attacks for different tasks.

\noindent\textbf{Adversarial defense.}
There have been several strategies for defense to eliminate threat of adversarial perturbation. 
A major class of defense transforms the input images for high robustness \cite{guo2017countering,xie2017mitigating}.
Such approaches translate pixel values of adversarial/clean samples to remove the influence of highly transferable adversarial perturbation.

Current input transformation based defenses that employ DGNs can be divided into three categories, according to their training constraints.
1) Using pixel-level constraints to reduce the differences of pixel values between clean and adversarial samples
\cite{shen2017ape,mustafa2019image,samangouei2018defense,prakash2018deflecting,song2017pixeldefend,huang2021advfilter};
2) applying feature-level constraints to unify representations of clean and adversarial samples in feature space of the target model \cite{liao2018defense}; 
3) simultaneously setting pixel- and feature-level constraints, which are proved to be more advantageous \cite{naseer2020self,li2020enhancing}. 
We note in the feature level, existing approaches only set the distance between clean and adversarial samples as the constraint to optimize, without alignment of distributions in feature space.
For example, Huang et.al.~\cite{huang2021advfilter} proposed to train a network with the mechanism of predictive perturbation-aware filtering, which can remove the adversarial perturbation through a denoising operation.
In this paper, we propose to implement a novel input transformation strategy, where the DGNs are trained with exact pixel-level and feature-level distribution alignment.

Moreover, besides the lack of pixel-level alignment, although some existing adversarial training approaches have considered the feature level alignment, our method has noticeable differences with them.
For instance, Mustafa et.al.~\cite{mustafa2019adversarial,mustafa2020deeply} designed prototype conformity loss which can force the features for each class to lie inside a convex polytope that is maximally separated from the polytopes of other classes, while such a loss can not achieve the distribution alignment for adversarial samples and clean samples in the deep feature space.
Hou et.al.~\cite{hou2020class} proposed to set a discriminator and adopt an adversarial learning strategy, which is similar to GAN training~\cite{goodfellow2014generative}, to make the features of adversarial samples and clean samples indistinguishable. However, such adversarial learning can not explicitly enforce the complete distribution alignment, leading to the misalignment in the deep feature space with high probability.
Song et.al.~\cite{song2019improving} proposed to incorporate the constraints of distribution alignment in the adversarial training. Nevertheless, such a method has not considered achieving an exact distribution alignment which is our target: not only the distribution shapes of adversarial and clean samples are aligned, but also the features of the paired individuality (the paired adversarial and the corresponding clean sample) are also aligned.

\begin{figure*}[t]
	\begin{center} 
		\subfigure[Visual illustration for traditional pixel-level training constraints]{\includegraphics[width=0.48\linewidth]{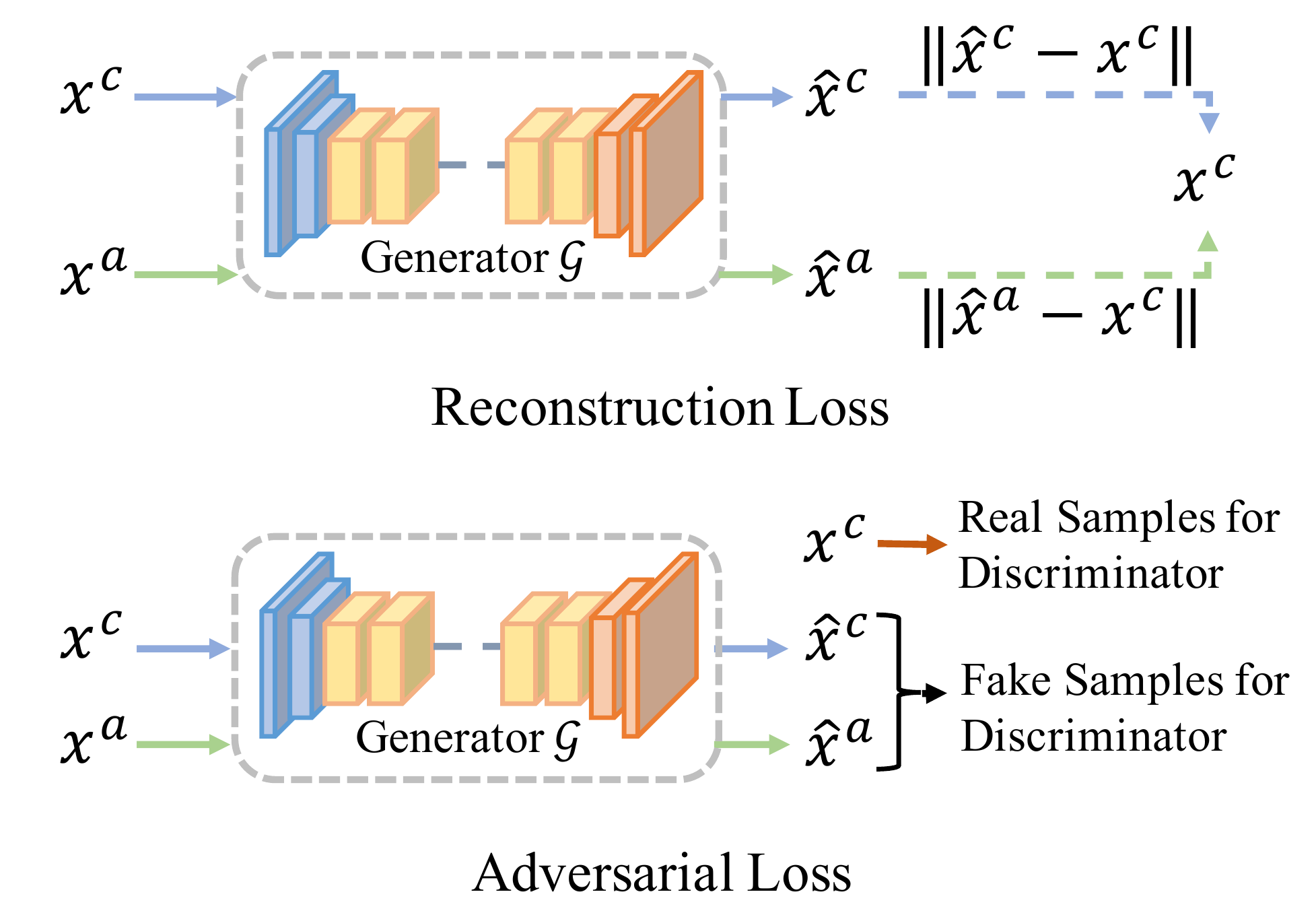}}
		\subfigure[Visual illustration for our pixel-level training constraints]{\includegraphics[width=0.48\linewidth]{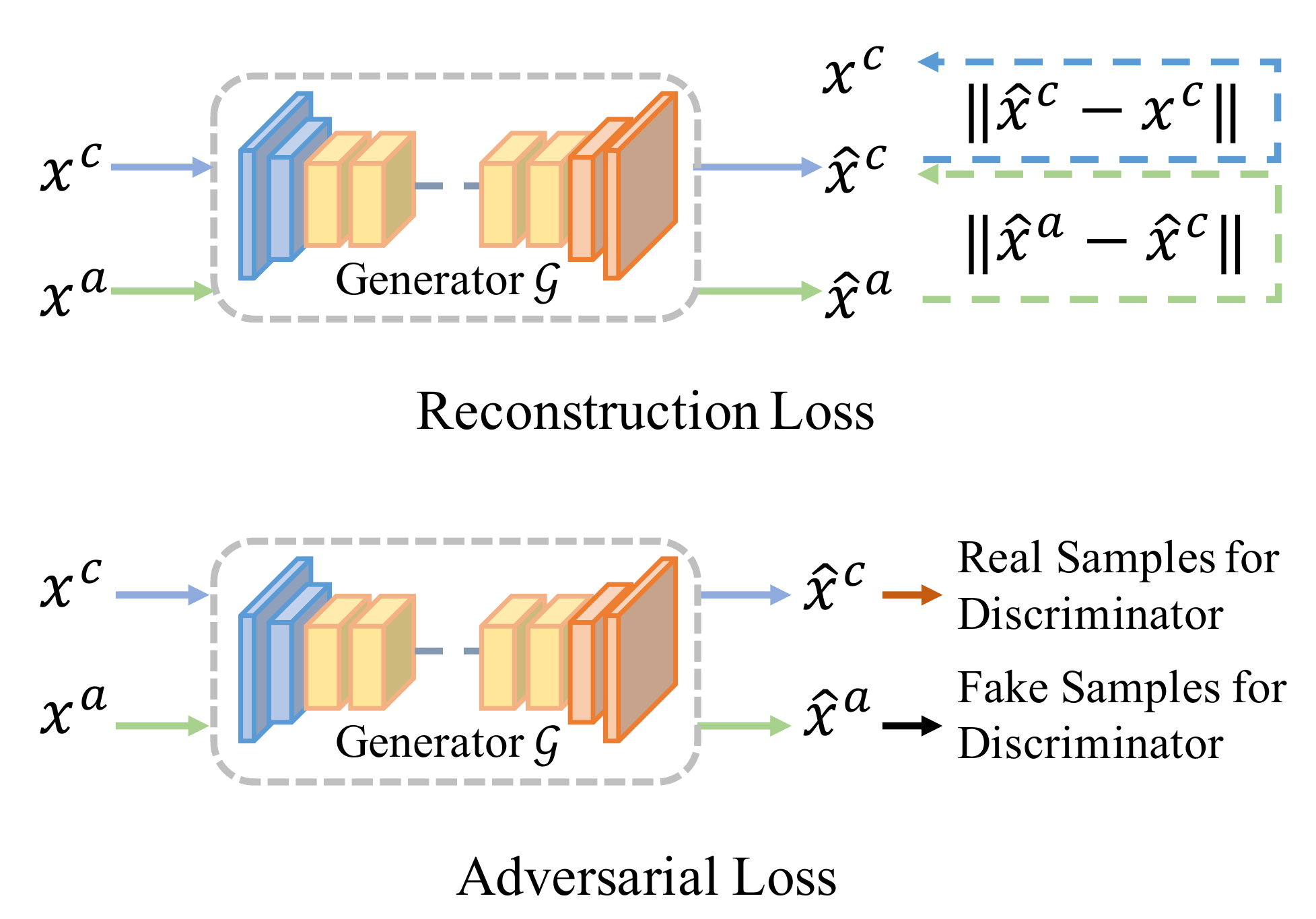}}
	\end{center}
	\vspace{-0.1in}
	\caption{Illustration for the differences between traditional and our pixel-level constraints.}
	\vspace{-0.1in}
	\label{fig:pix}
\end{figure*}

\section{Our Method}
We train a network $\mathcal{O}$ for one task $\mathcal{S}$, where $\mathcal{S}$ can be image classification, semantic segmentation, or object detection. The network $\mathcal{O}$ is usually trained with a set of clean samples $x^{c}$ and we suppose $x^{c} \sim \mathcal{C}$, where $\mathcal{C}$ is the distribution of clean samples.
Trained network $\mathcal{O}$ behaves decently on $x^{c} \sim \mathcal{C}$ for task $\mathcal{S}$, while its performance remarkably degrades after adding adversarial perturbation $\epsilon$ to $x^{c}$.

Adversarial samples are denoted as $x^{a}$ ($x^{a} = x^{c} + \epsilon$), and we represent the distribution of adversarial samples as $\mathcal{A}$.
As shown in Fig. \ref{fig:short}(a)\&(b), although the clean samples $x^{c}$ and the adversarial samples $x^{a}$ are indistinguishable, there is a large gap between $\mathcal{C}$ and $\mathcal{A}$ in feature space of the target model $\mathcal{O}$ that causes $\mathcal{O}$ to fail on adversarial samples. 

To eliminate the threat of adversarial samples, we align $\mathcal{C}$ and $\mathcal{A}$ for a target model $\mathcal{O}$ as exhibited in Fig. \ref{fig:short}(c)\&(d).
Based on this motivation, we propose to train a network $\mathcal{G}$ that aligns $\mathcal{C}$ and $\mathcal{A}$ in the feature space of the target model by modifying the pixel values of $x^{c}$ and $x^{a}$.

As a result, $\mathcal{G}$ translates $x^{c}$ into $\widehat{x}^{c}$, $x^{a}$ into $\widehat{x}^{a}$, and $\mathcal{O}$ achieves decent results on $\widehat{x}^{c}$ and $\widehat{x}^{a}$.
To train $\mathcal{G}$, we set constraints for alignment in both pixel and feature level as shown in Fig. \ref{fig:framework}, and push the distribution of $\widehat{x}^{c}$ and $\widehat{x}^{a}$ to approach $\mathcal{C}$ since $\mathcal{O}$ performs well on $x^{c} \sim \mathcal{C}$.
Furthermore, it is proved in Sec.~\ref{sec:eva_transfer} that the trained $\mathcal{G}$ can also be generalized to protect the models $\mathcal{O}'$ that have not appeared during training.

\subsection{Pixel-level Alignment}
\noindent\textbf{Motivation.}
The pixel-level training constraints are employed to weaken the difference in terms of pixel RGB values between clean and adversarial samples, and promote alignment in the feature space of the target model $\mathcal{O}$.
Previous work validated that ``distance metric between clean and adversarial samples'' and ``adversarial learning" \cite{goodfellow2014generative} are two practical pixel-level training constraints.
Traditional pixel-level constraints \cite{shen2017ape,mustafa2019image,samangouei2018defense,prakash2018deflecting,song2017pixeldefend} mainly adopt $x^c$ to guide formulation of $\widehat{x}^c$ and $\widehat{x}^a$ through the generator $\mathcal{G}$, as displayed in Fig. \ref{fig:pix}(a). They compute the distance metric between $\widehat{x}^c$ and $x^c$, $\widehat{x}^a$ and $x^c$. And they set $x^c$ as real samples, $\widehat{x}^c$ and $\widehat{x}^a$ as fake samples to conduct adversarial learning.
%%%Although this setting could result in reasonable performance on clean samples for $\mathcal{O}$, it has the chance to lead to unsatisfactory outcome on adversarial samples, primarily due to the ill-posed discrepancy between $\mathcal{C}$ and $\mathcal{A}$ -- it causes difficulty when directly matching $\widehat{x}^a$ to $x^c$ (see ``Theoretical analysis" for more details).

We propose a scheme for pixel-level training constraints, where we use $x^c$ to guide the formulation of $\widehat{x}^c$ and utilize $\widehat{x}^c$ to help matching $x^a$, as shown in Fig. \ref{fig:pix}(b). In this setting, $\widehat{x}^c$ is intermediate to shorten the discrepancy between $\widehat{x}^a$ and $x^c$ effectively. 
Comprehensive experiments illustrate that our novel setting results in conspicuous improvement for performance on adversarial samples, compared to the traditional schemes.

Our pixel-level training constraints also include the distance metric as well as adversarial learning.

\noindent\textbf{Reconstruction loss.} Given clean samples $x^{c}$, adversarial samples $x^{a}$ are obtained by adding adversarial perturbation $\epsilon$ to $x^{c}$, and the generator $\mathcal{G}$ synthesizes output $\widehat{x}^c$ and $\widehat{x}^a$ with input of $x^{c}$ and $x^{a}$. Based on it, we define the reconstruction loss term $\mathcal{L}_{{r}}$ as
\begin{equation}
\begin{aligned}
&\widehat{x}^a=\mathcal{G}(x^a), \ \widehat{x}^c=\mathcal{G}(x^c), \\ &\mathcal{L}_{{r}}=\mathbb{E}(\Vert \widehat{x}^c-x^c \Vert_1) + \mathbb{E}(\Vert \widehat{x}^a-\widehat{x}^c \Vert_1),\\
\end{aligned}
\label{pixel-rec}
\end{equation}
where $\mathbb{E}$ is to compute the mean value, $\Vert \, \Vert_1$ is the $L_1$ Euclidean distance.
Moreover, to help synthesize images with high resolutions,
we use a perceptual loss term \cite{johnson2016perceptual,zhu2017unpaired} $\mathcal{L}_{{p}}$, by computing the reconstruction distance in the feature space of an ImageNet-pretrained VGG-16 network \cite{simonyan2014very} for $\widehat{x}^c$ and $x^c$, as well as $\widehat{x}^a$ and $\widehat{x}^c$.
Not that the perceptual loss term is not the constraint for feature-level alignment, since the VGG-16 network is not task-specific and we mainly employ the perceptual loss for the visual similarity of $\widehat{x}^c$ and $x^c$ (or $\widehat{x}^a$ and $\widehat{x}^c$) in the pixel level.

\noindent\textbf{Adversarial loss.} 
To match the global pattern for $\mathcal{C}$ and $\mathcal{A}$ at the pixel level, 
we implement loss terms in adversarial learning with sampling. We employ a discriminator $\mathcal{D}$, and set the loss terms in the form of LSGAN \cite{mao2017least} as
\begin{equation}
\begin{aligned}
\mathcal{L}_{{GAN}_{d}} = &\mathbb{E}_{x^c \sim \mathcal{C}} ((\mathcal{D}(\widehat{x}^c)-1)^2) +\\& \mathbb{E}_{x^a \sim \mathcal{A}} ((\mathcal{D}(\widehat{x}^a)-0)^2), \\ \mathcal{L}_{{GAN}_{g}} = &\mathbb{E}_{x^a \sim \mathcal{A}}((\mathcal{D}(\widehat{x}^a)-1)^2),
\end{aligned}
\label{gan1}
\end{equation}
where $\mathcal{L}_{{GAN}_{d}}$ is set for the discriminator, and $\mathcal{L}_{{GAN}_{g}}$ is adopted for the generator.
Additionally, the feature match loss is adopted as an auxiliary part of the adversarial loss \cite{wang2018high}. We obtain intermediate features from $\mathcal{D}$ for fake and real samples, and compute their distances as
\begin{equation}
\begin{split}
\mathcal{L}_{{m}} &=\mathbb{E}(\Vert \mathcal{F}(\widehat{x}^c) -\mathcal{F}(\widehat{x}^a) \Vert_1),
\end{split}
\label{fm1}
\end{equation}
where $\mathcal{F}(x)$ are intermediate features obtained from the discriminator for real or fake samples $x$.

\begin{algorithm*}[t]
	\caption{Our strategy to train the generator $\mathcal{G}$ for the target model $\mathcal{O}$ on the target task $\mathcal{S}$}
	\label{alg:train}
	{\bf Parameter:} 
	Training data $(x^c, y^c)$, initialized $\mathcal{G}$ and $\mathcal{D}$, maximum number of iteration $T_{max}$, number of iteration $T \leftarrow 0$
	\begin{algorithmic}[1]
		\While{$T\not= T_{max}$}
		\State {Read a minibatch of data $D_b = \{x^c_1, ..., x^c_b\}$, $Y_b = \{y^c_1, ..., y^c_b\}$.}
		\State {Use the chosen attack algorithm and $\mathcal{O}$ to generate adversarial samples $A_b = \{x^a_1, ..., x^a_b\}$.}
		\State {Compute $\mathcal{L}_{{r}}$, $\mathcal{L}_{{p}}$, $\mathcal{L}_{{GAN}_{g}}$ and $\mathcal{L}_{{m}} $, using $D_b$, $A_b$ and the discriminator $\mathcal{D}$.}
		\State {Forward $D_b$ and $A_b$ through $\mathcal{O}$ to obtain features of $K$ classes, according to $Y_b$.}
		\State {Compute the clustering center of each class according to the obtained features in this batch.}
		\State {Compute $\mathcal{L}_{{F}_{class}}$ using features of $K$ classes and the corresponding clustering centers.}
		\State{Compute $\mathcal{L}_{g}$ to update the generator $\mathcal{G}$, compute $\mathcal{L}_{{GAN}_{d}}$ to update the discriminator $\mathcal{D}$. $T \leftarrow T + 1$}
		%\State{$T \leftarrow T + 1$.}
		\EndWhile
	\end{algorithmic}
\end{algorithm*}

\noindent\textbf{Theoretical analysis.} In the traditional setting, to weaken the discrepancy between adversarial samples with the corresponding clean samples, $\Vert \widehat{x}^a-x^c\Vert_1$ is set as the objective to optimize for training $\mathcal{G}$.
However, $\Vert \widehat{x}^a-x^c\Vert_1$ is large at the beginning of training and there exist multiple solutions $\widehat{x}^a$ that have the same value for $\Vert \widehat{x}^a-x^c\Vert_1$ at each step of training. 
Thus, this setting is ill-posed during training and is very likely to result in local optima (i.e., $\widehat{x}^a$ generated from $\mathcal{G}$ does not approach $x^c$ enough).

\begin{figure}[t]
	\begin{center} 
		\includegraphics[width=1.0\linewidth]{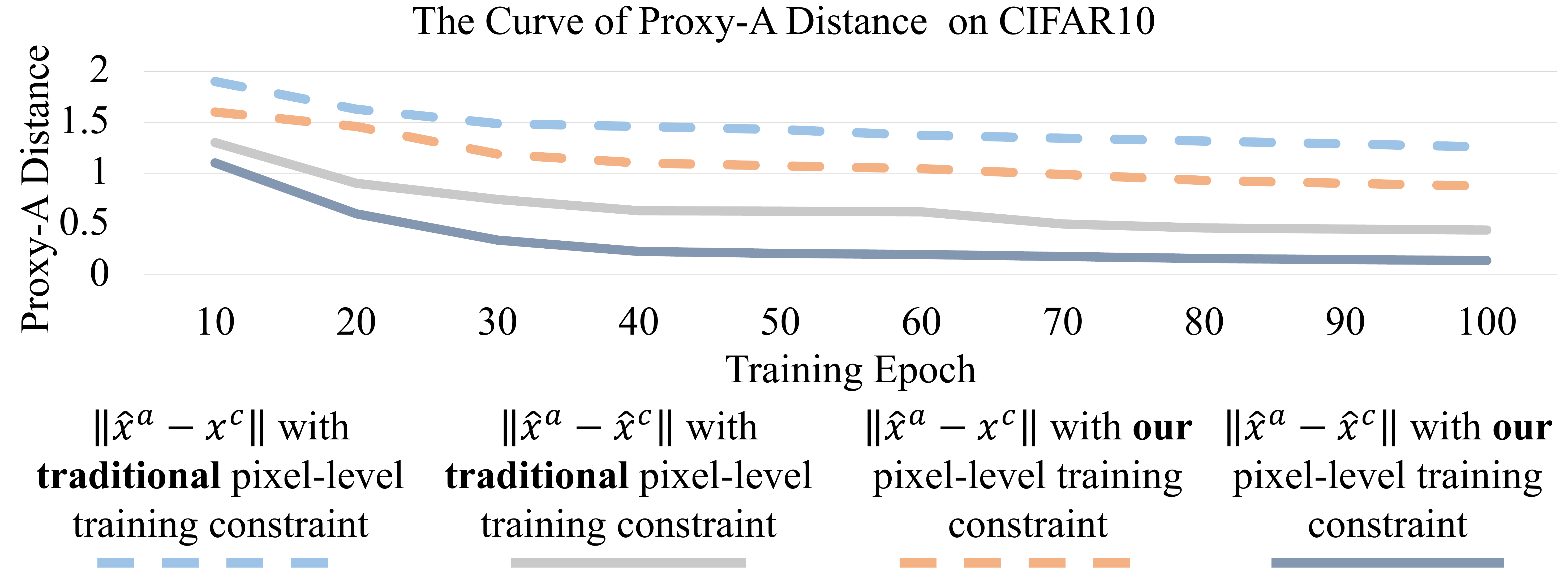}
	\end{center}
	\vspace{-0.2in}
	\caption{The visualization of Proxy-A distance between ``$\widehat{x}^a$ and $x^c$" as well as ``$\widehat{x}^a$ and $\widehat{x}^c$" at each training epoch. 
	With our pixel-level training constraint, $\widehat{x}^a$ can be better aligned with $\widehat{x}^c$ and $x^c$.}
	\vspace{-0.1in}
	\label{fig:pix22}
\end{figure}

In our pixel-level constraint, we instead set $\Vert \widehat{x}^a-\widehat{x}^c\Vert_1$ as the objective. The distance between $\widehat{x}^a$ and $\widehat{x}^c$ is narrower than the distance between $\widehat{x}^a$ and $x^c$, since $\widehat{x}^a$ and $\widehat{x}^c$ locate in the same output space of $\mathcal{G}$. 
To demonstrate this, we adopt Proxy-A distance \cite{ben2006analysis} to measure the distance between two domains' distributions.
Given the generalization error $\kappa$ on discriminating between the target and source samples, the Proxy-A distance is defined as $2(1-2\kappa)$.
It is observed that the Proxy-A distance between $\widehat{x}^a$ and $\widehat{x}^c$ is smaller at the beginning of training, always shorter than the Proxy-A distance between $\widehat{x}^a$ and $x^c$, as shown in Fig.~\ref{fig:pix22}. It finally approaches zero.

%%\xgxu{We provide a visualization in Fig. which represents the distance between $\widehat{x}^a$ and $x^c$, as well as the distance between $\widehat{x}^a$ and $\widehat{x}^c$, in the training process of traditional strategy and our strategy.
%%Obviously, the distance between $\widehat{x}^a$ and $\widehat{x}^c$ is always shorter than the distance between $\widehat{x}^a$ and $x^c$, and such distances can be decreased faster and lower in our strategy.}

Thus, our setting has a smaller solution space during training and is decent to avoid local optima, since it is simpler to train $\mathcal{G}$ to make $\widehat{x}^a$ very close to $\widehat{x}^c$. And we find that the optimization between $\widehat{x}^a$ and $\widehat{x}^c$ can also lead to more narrow distance between $\widehat{x}^a$ and $x^c$.
It is verified by the visualizations in Fig. \ref{fig:short} and Fig. \ref{fig:pix22}.

Moreover, the target model $\mathcal{O}$ performs well on both $x^c$ and $\widehat{x}^c$, validated in experiments. $\widehat{x}^a$ obtained from our setting is close to $x^c$ because $\widehat{x}^c$ is close to $x^c$, and $\widehat{x}^a$ is very near $\widehat{x}^c$. Therefore, $\mathcal{O}$ behaves better on $\widehat{x}^a$ with our setting than traditional ones.

\subsection{Feature-level Alignment}
\noindent\textbf{Motivation.}
Besides the pixel-level training constraints, recent work revealed the necessity of feature-level training constraints for the target model $\mathcal{O}$ \cite{liao2018defense,naseer2020self} to enhance protection.
Feature-level alignment helps formulate robust representation for networks of a given task.
Existing methods formulate feature-level training constraints as the distances in feature space of $\mathcal{O}$ between synthesized and clean samples.
However, they ignore the constraint for the alignment of overall $\mathcal{C}$ and $\mathcal{A}$ in feature level.
We instead propose a class-aware feature-level training constraint, besides the distance metric in the feature space of $\mathcal{O}$. 
Our class-aware constraint aligns the integrated distribution of clean and adversarial samples within each category for the target model, minimizing the intra-class distance and maximizing the inter-class distance for adversarial samples' distribution.

\noindent\textbf{Task-oriented loss.}
Suppose $\mathcal{O}$ is trained with the paired data $(x^c, y^c)$ where $y^c$ is the ground truth for $x^c$. And the loss term to train $\mathcal{O}$ can be represented as $\mathcal{L}_{{o}} (x^c, y^c)$ (e.g., $\mathcal{L}_{{o}}$ is the cross-entropy loss for image classification and semantic segmentation, the loss for bounding box regression and classification in the object detection task).
To align behaviors of clean and adversarial samples in feature space of $\mathcal{O}$, we adopt a feature-level loss as
\begin{equation}
\begin{split}
\mathcal{L}_{F_{task}} &=\mathcal{L}_{{o}} (\widehat{x}^c, y^c) + \mathcal{L}_{{o}} (\widehat{x}^a, y^c).
\end{split}
\label{feature-cls}
\end{equation}

\noindent\textbf{Reconstruction loss.}
Similar to the reconstruction loss in pixel level, we set loss $\mathcal{L}_{{F}_{rec}}$ to minimize distances between adversarial and clean samples in the feature space of $\mathcal{O}$ as
\begin{equation}
\begin{aligned}
&\widehat{z}^a=\mathcal{O}(\widehat{x}^a), \ \widehat{z}^c=\mathcal{O}(\widehat{x}^c), \ z^c=\mathcal{O}(x^c), \\ &\mathcal{L}_{{F}_{rec}}=\mathbb{E}(\Vert \widehat{z}^c-z^c \Vert_1) + \mathbb{E}(\Vert \widehat{z}^a-z^c \Vert_1),\\
\end{aligned}
\label{feature-rec}
\end{equation}
where $\widehat{z}^a$, $\widehat{z}^c$ and $z^c$ are the intermediate features of $\widehat{x}^a$, $\widehat{x}^c$ and $x^c$ in the feature space of $\mathcal{O}$. The choice of feature space to compute feature-level constraints for different tasks is illustrated in the supplementary file.

\noindent\textbf{Distribution alignment loss.}
We also note previous methods do not consider aligning the overall distribution of $\mathcal{C}$ and $\mathcal{A}$ in the feature space of $\mathcal{O}$. In contrast, we formulate our class-aware training constraint for the overall distribution alignment.
Suppose there are $K$ classes within the distribution of $\mathcal{C}$ and $\mathcal{A}$. We represent $z^{c(k)}$ and $\widehat{z}^{a(k)}$ as the features for $x^c$ and $\widehat{x}^a$ of the $k$-th class extracted from $\mathcal{O}$.
The clustering center of $z^{c(k)}$ and $\widehat{z}^{a(k)}$ is denoted as $m^{c(k)}$ and $\widehat{m}^{a(k)}$ in the distribution of $\mathcal{C}$ and $\mathcal{A}$ respectively.
The class-aware constraint consists of three terms. First, we compute a loss term as the distance between $m^{c(k)}$ and $\widehat{m}^{a(k)}$ to align the distribution of adversarial samples to clean samples as
\begin{equation}
\begin{aligned}
\mathcal{L}_{{F}_{align}}&=\sum_{k=1:K} \mathbb{E}(\Vert m^{c(k)}-\widehat{m}^{a(k)}\Vert_1).\\
\end{aligned}
\label{feature-align}
\end{equation}
Further, favorable distributions in the feature spaces of the target models for high-level tasks should have wide distances between the features from different classes, and narrow separation among the features from the same class. To this end, we set intra- and inter-class loss as
\begin{equation}
\begin{aligned}
&\mathcal{L}_{{F}_{intra}}=\sum_{k=1:K} \mathbb{E}(\Vert \widehat{z}^{a(k)}- \widehat{m}^{a(k)} \Vert_1), \\ &\mathcal{L}_{{F}_{inter}}=\sum_{k=1:K} \sum_{i=1:K, i\neq k} \mathbb{E}(M-\Vert \widehat{m}^{a(k)}-\widehat{m}^{a(i)}\Vert_1),\\
\end{aligned}
\label{feature-intra}
\end{equation}
where $M$ is a pre-defined hyper-parameter for controlling the inter-class distance. And the overall class-aware training constraint is written as
\begin{equation}
\begin{aligned}
\mathcal{L}_{{F}_{class}}&=\lambda_1 \mathcal{L}_{{F}_{align}}+\lambda_2 \mathcal{L}_{{F}_{inter}}+\lambda_3 \mathcal{L}_{{F}_{intra}},\\
\end{aligned}
\label{feature-class}
\end{equation}
where $\lambda_1$, $\lambda_2$, and $\lambda_3$ are loss weights. 

\subsection{Overall Training Constraint}
In summary, the overall training constraint for the generator $\mathcal{G}$ can be written as
\begin{equation}
\begin{aligned}
\mathcal{L}_{{g}}=&\lambda_4 (\mathcal{L}_{{r}}+\mathcal{L}_{{p}}+\mathcal{L}_{{m}}+\mathcal{L}_{{GAN}_{g}} + \mathcal{L}_{F_{task}} + \mathcal{L}_{{F}_{rec}})+\\&\lambda_1 \mathcal{L}_{{F}_{align}}+\lambda_2 \mathcal{L}_{{F}_{inter}}+\lambda_3 \mathcal{L}_{{F}_{intra}},\\
\end{aligned}
\label{loss_over}
\end{equation}
where $\lambda_1$ to $\lambda_4$ are loss weights.
Our training algorithm for $\mathcal{G}$ is summarized in Alg. \ref{alg:train}.
$\lambda_1$, $\lambda_2$, $\lambda_3$ and $\lambda_4$ are set using grid search on the validation set.
We adopt $\lambda_1=0.1$, $\lambda_2=1$, $\lambda_3=0.005$ and $\lambda_4=50$ in our experiments.
We apply the same loss weights for training of various tasks.

Note that {adversarial samples $x^a$ during training are synthesized from the target model $\mathcal{O}$ only, while the trained model $\mathcal{G}$ can defend unseen adversarial perturbations that are transferred from models with different structures}.

\section{Experiments}
%\vspace{-0.05in}
Our framework is applicable to several important tasks. In the following, we conduct extensive experimental evaluation on image classification, semantic segmentation and object detection, on multiple datasets.
We show the effect of our proposed pixel- and feature-level constraints through ablation study and illustrate the superiority of our method compared with other alternatives.

\vspace{-0.05in}
\subsection{Datasets}
To demonstrate the performance of our method on the three key tasks, we select representative datasets in each task.
For image classification, we choose CIFAR10 \cite{krizhevsky2009learning}, CIFAR100 \cite{krizhevsky2009learning} and ImageNet \cite{deng2009imagenet} (we adopt its subset, the Tiny-ImageNet with 200 classes); for semantic segmentation, we employ Cityscapes \cite{cordts2016cityscapes} and VOC2012 \cite{pascal-voc-2012}; for object detection, VOC07+12 \cite{pascal-voc-2012} setting is adopted. The train and test splits are set according to their official setting.

\vspace{-0.05in}
\subsection{Structure of DCNs in Experiments}
In our experiments, the generator $\mathcal{G}$ is implemented as the ``global generator" structure in pix2pixHD \cite{wang2018high}.
In classification, images from CIFAR10 and CIFAR100 are with size $32 \times 32$ and the images from Tiny-ImageNet are $64 \times 64$ large for the input of target model as well as the generator. The generator contains two down-sample/up-sample layers.
In the semantic segmentation task, the images from Cityscapes and VOC2012 are shaped as $512 \times 512$ for the generator with four down-sample/up-sample layers, and are set as $425 \times 425$ and $417 \times 417$ for the target model respectively.
As for object detection, the input size of the generator, which is built with four down-sample/up-sample layers, is $256 \times 256$ for VOC07+12. The input size of the target model is $300 \times 300$.
Note that our training constraints are suitable for the generator $\mathcal{G}$ that is built with various structures.

\vspace{-0.05in}
\subsection{Target Models}
In the classification task, $\mathcal{O}$ is adopted as the structure of WideResNet \cite{zhang2019theoretically}/ResNet50 \cite{he2016deep}; as for semantic segmentation, we employ $\mathcal{O}$ with the architecture of PSPNet \cite{zhao2017pyramid}/DeepLabv3 \cite{chen2017rethinking}; in object detection, $\mathcal{O}$ is set as the framework of SSD \cite{liu2016ssd}/RFBNet \cite{liu2018receptive}.
We use $\mathcal{O}$ (trained without defense) to train $\mathcal{G}$ in input transformation strategies (including ours), and the results of processed clean/adversarial samples $\widehat{x}^c$/$\widehat{x}^a$ are computed by $\mathcal{O}$ during evaluation.
And in Sec.~\ref{sec:eva_transfer}, it is verified that $\mathcal{G}$ can also protect target models $\mathcal{O}'$ that have different structures and parameters from $\mathcal{O}$, and have not appeared during training.

\begin{table*}[t]
	\centering
	\caption{Comparison on the classification tasks. 
		``WideResNet $\rightarrow$ ResNet50": attacking WideResNet while generating adversarial samples from ResNet50.} 
	\vspace{-0.1in}
	\Huge
	\label{tab:black-box0}
	\resizebox{1.0\linewidth}{!}{
		\begin{tabular}{l|p{3.5cm}<{\centering}p{3.5cm}<{\centering}p{3.5cm}<{\centering}p{3.5cm}<{\centering}|p{3.5cm}<{\centering}p{3.5cm}<{\centering}p{3.5cm}<{\centering}p{3.5cm}<{\centering}|p{3.5cm}<{\centering}p{3.5cm}<{\centering}p{3.5cm}<{\centering}p{3.5cm}<{\centering}}
			\toprule[1pt]
			\multirow{2}{5cm}{WideResNet $\rightarrow$ ResNet50}&\multicolumn{4}{c|}{CIFAR10 (Accuracy \%)} &\multicolumn{4}{c|}{CIFAR100 (Accuracy \%)} &\multicolumn{4}{c}{Tiny-ImageNet (Accuracy \%)}  \\
			\cline{2-13}
			& {clean} & {PGD} & {DeepFool}& {C\&W} & {clean} & {PGD} & {DeepFool}& {C\&W} & {clean} & {PGD} & {DeepFool}& {C\&W}   \\
			\hline
			No Defense & 95.1&2.1&5.3&6.4&78.1&7.5&9.3&10.4&\textbf{64.5}&19.2&20.4 &21.2  \\
			No Defense (finetune) & \textbf{95.6} &3.8 &7.5&8.7 &\textbf{79.5}&8.0& 9.7&10.9&63.9&22.1&23.2&23.6 \\
			\hline
			TRADES \cite{zhang2019theoretically} &87.3  &79.7 &87.1& 87.2&62.8&53.9&62.6 &62.7&58.5&45.3&58.3&58.5 \\
			TRADES (finetune) \cite{zhang2019theoretically}  & 85.4 &78.5 &85.3& 85.4&78.1&53.9&55.8 &44.2&43.3&42.2&43.2&43.4 \\
			Free-adv \cite{shafahi2019adversarial} & 77.1 &73.1 &76.7&77.0 &49.2&46.6&49.0 &49.2&55.5&45.8&55.4&55.5 \\
			Free-adv (finetune) \cite{shafahi2019adversarial} & 88.5 &79.8 &88.4&88.5 &63.7&\textbf{54.1}&63.5 &63.7&42.2&40.9&48.6&48.7 \\
			\hline
			Defense \cite{samangouei2018defense} &39.9&38.7&38.1&39.3& 31.1 &30.5 &30.9& 31.0&20.4&18.5&19.7 &19.9 \\
			SR \cite{mustafa2019image}&48.0&47.2&48.0&48.1&  33.6& 33.1&33.5& 33.8&31.1&30.5&30.9 &31.2 \\
			FPD \cite{li2020enhancing}&48.5&47.6&48.4&48.5& 52.5 &41.2 &42.4&42.5 &39.7&33.7&39.5 &39.8 \\
			APE \cite{shen2017ape} &90.2&41.9&89.1&89.8& 73.2 &37.2 &65.7& 67.7&62.4&30.5&61.6 &62.3 \\
			Denoise \cite{liao2018defense}&89.8&80.0&90.1&90.0& 67.2 &53.1 &66.8&66.1 &59.8&45.3&59.4 &59.9 \\
			NRP \cite{naseer2020self} &91.8&79.0&89.2&89.8& 70.3 &52.8 &66.6&67.2 &59.1&45.5&59.0 &59.2 \\
			Ours &90.7&\textbf{81.4}&\textbf{90.4}&\textbf{90.6}&68.7&\textbf{54.0}&\textbf{67.5}&\textbf{68.4}&62.9&\textbf{46.4} &\textbf{62.2} &\textbf{63.2}\\
			\bottomrule[1pt]
	\end{tabular}}
\vspace{-0.1in}
\end{table*}

\vspace{-0.1in}
\subsection{Training}
\noindent\textbf{Training parameters.}
We employ Adam optimizer with $\beta_1$ and $\beta_2$ set to 0.5 and 0.999.
%, respectively. 
The learning rate is set as 0.0002 and the batch size is 1 for Cityscapes and VOC2012, 4 for VOC07+12 , 16 for Tiny-ImageNet, and 64 for CIFAR10/CIFAR100.
The number of training epoch is 80 for Cityscapes, 20 for VOC2012, 80 for VOC07+12, 30 for Tiny-ImageNet, 100 for CIFAR10/CIFAR100.
Our method is implemented with PyTorch \cite{paszke2019pytorch}, and runs on a TITAN X GPU. 

\vspace{0.05in}
\noindent\textbf{Data augmentation.}
For training on three tasks, we adopt the data augmentation policies described in \cite{zhang2019theoretically,zhao2017pyramid,liu2016ssd}. For CIFAR10 and CIFAR100, we first pad the input to $40\times 40$ and randomly crop a patch of size $32\times 32$. Then the patch is chosen to be horizontally flipped or not; for Tiny-ImageNet, the data augmentation includes the operation of random crop (the pad size is 4 and the crop size is 64) and random horizontal flip. 
For the segmentation task (Cityscapes and VOC2012), we employ the official augmentations in \cite{semseg2019}, including RandScale, RandRotate, RandomGaussianBlur, RandomHorizontalFlip, and random crop. 
For the detection task (VOC07+12), the augmentation strategies in \cite{liu2016ssd} are utilized.

\vspace{0.05in}
\noindent\textbf{Details of our training constraints for image classification.} For the classification task, we adopt WideResNet \cite{zhang2019theoretically} and ResNet50 \cite{he2016deep} for experiments, and implement feature-level training constraints by extracting the feature of the fully connected layer before the final layer.
Especially, each image $x \in \mathbb{R} ^{H\times W \times 3}$ ($H$ and $W$ are the height and width of the image) has one class label $y \in \mathbb{R} ^ K$ and one feature vector $f \in \mathbb{R} ^ L$ ($L$ is the length of the vector). And we group features into $K$ classes according to $y$.

\vspace{0.05in}
\noindent\textbf{Details of our training constraints for segmentation.} In the semantic segmentation task, we use PSPNet \cite{zhao2017pyramid} and DeepLabv3 \cite{chen2017rethinking} with ResNet50 as the backbone for experiments, and complete feature-level constraints by using the feature of the convolution layer before the final layer. Different with the classification task, each image in semantic segmentation task has multiple class labels and thus has multiple feature vectors with different classes: for an image $x \in \mathbb{R} ^{H\times W \times 3}$, it has the corresponding segmentation map $y \in \mathbb{R} ^{H\times W \times K}$. And the feature of the image can be denoted as $f' \in \mathbb{R} ^{h\times w \times L}$ with a resized segmentation map $y' \in \mathbb{R} ^{h\times w \times K}$. Then we can obtain a set of feature vectors with their corresponding labels as $\{ y_1,...,y_{h\times w} \}, y_i \in \mathbb{R} ^ K$,  $\{ f_1, ..., f_{h\times w} \}, f_i \in \mathbb{R} ^ L$, which are reshaped from $y'$ as well as $f'$. Such set of feature vectors can be grouped into $K$ classes and utilized in feature-level constraints.

\vspace{0.05in}
\noindent\textbf{Details of our training constraints for object detection.} We employ SSD \cite{liu2016ssd} and RFBNet \cite{liu2018receptive} with VGG16 as the backbone for experiments within the object detection task. And the features for feature-level constraints are obtained as the outputs of the backbone and are cropped with the ground truth of the bounding box. In this task, each image has multiple class labels for bounding boxes in this image, and thus also has multiple feature vectors with different classes: an image $x \in \mathbb{R} ^{H\times W \times 3}$ can have $B$ bounding boxes and one class label for each box. The feature of the image can be denoted as $f' \in \mathbb{R} ^{h\times w \times L}$ and we can obtain the feature of each bounding box on this feature map with the corresponding class label, as $\{ y_1,...,y_{B} \}, y_i \in \mathbb{R} ^ K$,  $\{ f_1, ..., f_{B} \}, f_i \in \mathbb{R} ^ L$.

%For experiments of classification, the adversarial perturbation $\epsilon$ during training is generated by PGD \cite{madry2017towards} with KL criterion; for semantic segmentation, we employ BIM  \cite{kurakin2016adversarial} for training; in object detection, classification attack (``cls") and localization attack (``loc") \cite{zhang2019towards} are utilized during training.

\begin{table*}[t]
	\centering
	\caption{Comparison with existing defense methods on the semantic segmentation and object detection tasks. 
		%The symbolic representation is the same as that of Table \ref{tab:black-box}. 
		``PSPNet $\rightarrow$ DeepLabv3": attacking PSPNet while obtaining adversarial samples with DeepLabv3; ``SSD $\rightarrow$ RFBNet": attacking SSD while acquiring adversarial samples from RFBNet.} 
	\vspace{-0.1in}
	\Huge
	\label{tab:black-box1}
	%@{\hspace{0.8cm}}
	\resizebox{1.0\linewidth}{!}{
		\begin{tabular}{l|p{3.5cm}<{\centering}p{3.5cm}<{\centering}p{3.5cm}<{\centering}p{3.5cm}<{\centering}|p{3.5cm}<{\centering}p{3.5cm}<{\centering}p{3.5cm}<{\centering}p{3.5cm}<{\centering}|p{3.5cm}<{\centering}p{3.5cm}<{\centering}p{3.5cm}<{\centering}p{3.5cm}<{\centering}}
			\toprule[1pt]
			PSPNet $\rightarrow$ DeepLabv3&\multicolumn{4}{c|}{Cityscapes (mIoU \%)} &\multicolumn{4}{c|}{VOC2012 (mIoU \%)} &\multicolumn{4}{c}{VOC07+12 (mAP \%)}  \\
			\cline{2-13}
			SSD $\rightarrow$ RFBNet& {clean} & {BIM} & {DeepFool}& {C\&W} & {clean} & {BIM} & {DeepFool}& {C\&W} & {clean} & {cls} & {loc} & {cls+loc}   \\
			\hline
			No Defense &\textbf{73.5}&3.6&38.6&12.5& \textbf{76.4}&11.1 &46.1&16.6  &72.5&17.9 &14.5 &15.0 \\
			No Defense (finetune) &73.3&3.6&37.1&12.5&76.3 &11.2&45.2 &15.4 &\textbf{73.6}&19.6&16.8 &16.1 \\
			\hline
			SAT \cite{xu2020dynamic}  &65.7&50.5&52.5&51.0&73.9 &57.1&64.3 &69.3  &--&--&-- &-- \\
			SAT (finetune) \cite{xu2020dynamic} &66.3&42.1&47.7&42.4&74.2 &60.5&60.1 &63.4  &--&--&-- &--\\
			DDCAT \cite{xu2020dynamic} &67.7&51.4&54.4&51.8& 75.1&58.8&65.1 &69.3  &--&--&-- &-- \\
			DDCAT  (finetune) \cite{xu2020dynamic} &68.3&43.3&49.4&43.1&76.0 &62.4&62.4 &64.8  &--&--&-- &--\\
			CLS \cite{zhang2019towards}  &--&--&--&--& --&--& --&  --&47.8&34.5&43.3 &44.1 \\
			LOC \cite{zhang2019towards} &--&--&--&--& --&--& --&  --&52.9&36.7&38.8 &39.9 \\
			CON \cite{zhang2019towards}&--&--&--&--&-- &--& --& -- &40.7&31.3&39.5 &40.3 \\
			MTD \cite{zhang2019towards} &--&--&--&--& --&--& --&  --&49.1&42.0&44.3 &44.1 \\
			\hline
			Defense \cite{samangouei2018defense}&22.2&20.2&19.9&21.2&23.5 &21.9&22.6 &23.3 &34.6&30.2&30.8 &29.7 \\
			{SR \cite{mustafa2019image}} &60.7&52.0&50.6&51.5& 71.2&58.1&60.3 &67.2 &54.6&36.2&37.5 &34.5 \\
			FPD \cite{li2020enhancing}&55.9&53.0&53.2&55.0& 61.5&57.5& 58.6&59.7 &57.2&58.1&55.8 &57.8 \\
			APE \cite{shen2017ape}&54.5&34.7&33.3&45.4& 74.3&37.8&54.3 &59.9 &62.3&57.9&58.0 &55.8 \\
			Denoise \cite{liao2018defense}&64.4&55.1&53.8&64.0&70.4 &61.9&61.5 &67.8 &61.6&52.2&50.9 &51.7 \\
			NRP \cite{naseer2020self}  &65.0&55.2&49.3&64.1&70.5 &62.6& 59.3& 68.5&60.4&59.9&55.4 &58.9 \\
			Ours &67.6&\textbf{59.5}&\textbf{62.0}&\textbf{64.7}&71.2 &\textbf{63.6} &\textbf{66.0}& \textbf{69.5} &60.5& \textbf{61.2}&\textbf{58.3} &\textbf{60.9} \\
			\bottomrule[1pt]
	\end{tabular}}
\vspace{-0.1in}
\end{table*}

\begin{table*}[t]
	%\centering
	%\caption{The comparison between our approach and existing methods on the classification task. ``ResNet50 $\rightarrow$ WideResNet" means attacking ResNet50 while generating adversarial perturbations from WideResNet.} 
	\caption{Comparison on the classification tasks. 
		``ResNet50 $\rightarrow$ WideResNet": attacking ResNet50 while generating adversarial samples from WideResNet.} 
	\vspace{-0.1in}
	\Huge
	\label{tab:black-box111}
	\resizebox{1.0\linewidth}{!}{
		\begin{tabular}{l|p{3.5cm}<{\centering}p{3.5cm}<{\centering}p{3.5cm}<{\centering}p{3.5cm}<{\centering}|p{3.5cm}<{\centering}p{3.5cm}<{\centering}p{3.5cm}<{\centering}p{3.5cm}<{\centering}|p{3.5cm}<{\centering}p{3.5cm}<{\centering}p{3.5cm}<{\centering}p{3.5cm}<{\centering}}
			\toprule[1pt]
			\multirow{2}{5cm}{ResNet50 $\rightarrow$ WideResNet}&\multicolumn{4}{c|}{CIFAR10 (Accuracy \%)} &\multicolumn{4}{c|}{CIFAR100 (Accuracy \%)} &\multicolumn{4}{c}{Tiny-ImageNet (Accuracy \%)}  \\
			\cline{2-13}
			& {clean} & {PGD} & {DeepFool}& {C\&W} & {clean} & {PGD} & {DeepFool}& {C\&W} & {clean} & {PGD} & {DeepFool}& {C\&W}   \\
			\hline
			{No Defense} &94.3&8.2&4.3&5.8&76.1&17.3&18.3&16.6&\textbf{62.1}&22.7&21.7 &22.2   \\
			{No Defense (finetune)} &\textbf{94.7}&8.1&6.6&7.5& \textbf{77.8} &16.6 &18.3&16.6 &61.3&22.7& 25.8&26.1 \\
			\hline
			{TRADES \cite{zhang2019theoretically}} &85.3&82.3&85.1&85.2&  59.8&57.3 &59.7&59.8 &44.3&43.0&44.3 &44.3 \\
			{TRADES (finetune) \cite{zhang2019theoretically}}  &83.2&82.0&83.1&83.2& 56.3 &55.2 &56.3&56.3 &43.3&42.4&43.3 &43.3 \\
			{Free-adv \cite{shafahi2019adversarial}} &86.3&81.6&86.2&86.3& 64.0 &55.8 &63.9& 64.0&53.4&41.6& 53.3&53.4 \\
			{Free-adv (finetune) \cite{shafahi2019adversarial}} &88.3&82.9&88.2&88.3& 63.4&55.2& 63.3& 63.4&39.9&38.5 &39.7&39.8 \\
			\hline
			{Defense \cite{samangouei2018defense}} &40.1&38.8&39.7&39.8& 31.1 &30.2 &30.5& 30.9&21.5&19.9&20.3 &21.0 \\
			{SR \cite{mustafa2019image}} &45.3&45.1&45.3&45.3&  35.6& 35.3&35.9&36.3 &30.6&29.5&30.0 &30.1 \\
			{FPD \cite{li2020enhancing}}&53.0&52.7&52.9&53.0& 50.0 &46.8 &49.7&50.0 &34.5&31.7&34.2 &34.4 \\
			{APE \cite{shen2017ape}} &89.1&64.9&88.9&89.6& 71.7 &51.6 &66.4&67.2 &60.2&35.1&59.0 &59.1 \\
			{Denoise \cite{liao2018defense}} &88.3&80.7&87.9&88.2& 66.7 &56.3 &65.3& 65.5&56.6&42.3&56.5 &56.7 \\
			{NRP \cite{naseer2020self}} &90.3&80.9&88.1&88.3& 66.7 &56.6 &66.2& 66.6&56.5&\textbf{43.3}&56.1 &56.5 \\
			Ours&90.0&\textbf{84.0}&\textbf{89.8}&\textbf{90.2}&67.3&\textbf{57.7}&\textbf{67.5}&\textbf{68.1}&61.0&\textbf{43.1}&\textbf{59.9} &\textbf{60.8}\\
			%%%%%%%%%%%%%%%%%%%%%%%%%%%%%%%%%%%%%%%
			\bottomrule[1pt]
	\end{tabular}}
\vspace{-0.1in}
\end{table*}

\begin{table*}[t]
	%\caption{The comparison between our approach and existing methods on the semantic segmentation and object detection tasks. ``DeepLabv3 $\rightarrow$ PSPNet" means attacking DeepLabv3 while generating adversarial perturbations from PSPNet; ``RFBNet $\rightarrow$ SSD" means attacking RFBNet while generating adversarial perturbations from SSD.} 
		\caption{Comparison with existing defense methods on the semantic segmentation and object detection tasks. 
		``DeepLabv3 $\rightarrow$ PSPNet": attacking DeepLabv3 while obtaining adversarial samples with PSPNet; ``RFBNet $\rightarrow$ SSD": attacking RFBNet while acquiring adversarial samples from SSD.} 
	\vspace{-0.1in}
	\Huge
	\label{tab:black-box444}
	\resizebox{1.0\linewidth}{!}{
		\begin{tabular}{l|p{3.5cm}<{\centering}p{3.5cm}<{\centering}p{3.5cm}<{\centering}p{3.5cm}<{\centering}|p{3.5cm}<{\centering}p{3.5cm}<{\centering}p{3.5cm}<{\centering}p{3.5cm}<{\centering}|p{3.5cm}<{\centering}p{3.5cm}<{\centering}p{3.5cm}<{\centering}p{3.5cm}<{\centering}}
			\toprule[1pt]
			DeepLabv3 $\rightarrow$ PSPNet&\multicolumn{4}{c|}{Cityscape (mIoU \%)} &\multicolumn{4}{c|}{VOC2012 (mIoU \%)} &\multicolumn{4}{c}{VOC07+12 (mAP \%)}  \\
			\cline{2-13}
			RFBNet $\rightarrow$ SSD& {clean} & {BIM} & {DeepFool}& {C\&W} & {clean} & {BIM} & {DeepFool}& {C\&W} & {clean} & {cls} & {loc} & {cls+loc}   \\
			\hline
			{No Defense} &73.2&3.8&32.3&13.5& \textbf{76.9}&11.8 &49.0& 20.2 &80.6&16.9 &15.6& 15.9 \\
			{No Defense (finetune)} &\textbf{73.5}&3.9&33.1&13.3&75.7 &11.7&49.8 &18.4 &\textbf{80.8}&17.2&15.8 &16.3 \\
			\hline
			{SAT \cite{xu2020dynamic}}  &64.3&49.2&50.7&49.9& 72.8&56.2&65.3 &68.6  &--&--&-- &-- \\
			{SAT (finetune) \cite{xu2020dynamic}} &65.4&42.5&48.5&42.8&74.8 &60.1&64.1 &69.9  &--&--&-- &--\\
			{DDCAT \cite{xu2020dynamic}} &67.7&50.3&52.9&50.8& 74.2&61.5&66.7 &69.9  &--&--&-- &-- \\
			{DDCAT  (finetune) \cite{xu2020dynamic}} &68.2&43.3&50.1&43.6& 76.2&63.4&65.4 &70.2  &--&--&-- &--\\
			{CLS \cite{zhang2019towards}}  &--&--&--&--& --&--& --&  --&52.0&39.6&48.8 &49.1 \\
			{LOC \cite{zhang2019towards}} &--&--&--&--& --&--& --&  --&57.6&41.7&42.8 &44.4 \\
			{CON \cite{zhang2019towards}}&--&--&--&--&-- &--& --& -- &43.5&35.3&43.8 &44.1 \\
			{MTD \cite{zhang2019towards}} &--&--&--&--& --&--& --&  --&53.7&47.1&48.3 &48.9 \\
			\hline
			{Defense \cite{samangouei2018defense}} &23.3&21.2&20.4&22.9&22.5 &20.7&22.0 &22.4 &42.2&37.8&38.2 &37.1 \\
			{SR \cite{mustafa2019image}} &62.8&51.6&50.6&52.8&70.1 &60.5& 64.0& 68.6&56.2&38.6&39.3 &36.2 \\
			{FPD \cite{li2020enhancing}}&49.5&49.6&50.0&50.8&62.1 &59.7& 60.8&61.7 &58.7&59.3& 57.3&59.2 \\
			{APE \cite{shen2017ape}} &54.2&27.2&22.7&35.6&75.5 &40.8&57.6 &63.4 &68.5&60.0&61.3 &58.8 \\
			{Denoise \cite{liao2018defense}} &64.6&58.5&55.0&64.0&67.2 &63.4&64.7 &68.8 &68.5&59.3&57.9 &58.4 \\
			{NRP \cite{naseer2020self}}  &65.6&58.0&50.7&64.1& 68.1&64.7&62.8 &69.2 &68.5&67.2&63.5 &67.2 \\
			Ours &67.3&\textbf{60.3}&\textbf{62.7}&\textbf{64.9}&69.1 &\textbf{65.2} &\textbf{66.9}&\textbf{70.7}  &70.2& \textbf{70.0}& \textbf{68.1}&\textbf{70.0} \\
			%%%%%%%%%%%%%%%%%%%%%%%%%%%%%%%%%%%%%%%
			\bottomrule[1pt]
	\end{tabular}}
\vspace{-0.1in}
\end{table*}

\vspace{0.05in}
\noindent\textbf{Attacks employed in training.}
In this paper, all attacks are conducted with $L_{\infty}$ constraint and untargeted form (except the experiments in Sec.~\ref{sec:targeted}).
For the image classification task, we follow~\cite{zhang2019theoretically} to set the attack parameters during training and testing. And we employ the attack parameters' setting in \cite{xu2020dynamic} and \cite{zhang2019towards} for the semantic segmentation and object detection task, respectively.
Therefore, for experiments of classification, the adversarial perturbation $\epsilon$ during training is generated by PGD \cite{madry2017towards} with KL criterion; for semantic segmentation, we employ BIM  \cite{kurakin2016adversarial} for training; in object detection, classification attack (``cls") and localization attack (``loc") \cite{zhang2019towards} are utilized during training.
And the corresponding details are described as the following.
For the classification task, we utilize PGD attack with KL criterion during training, and the perturbation range $\epsilon=0.031 \times 255$, the step size $\alpha=0.0175\times 255$, the number of attack iteration $n=4$.
In the semantic segmentation task, we use BIM ($\epsilon=0.03 \times 255$, $\alpha=0.01\times 255$, $n=3$) during training.
Moreover, the attacks for classification loss and location loss are employed within the object detection task, and the perturbation range is 8 for pixel values within $[0, 255]$.

\vspace{-0.05in}
\subsection{Evaluation}
\noindent\textbf{Attack types.} During evaluation, in classification tasks, the adversarial perturbation is obtained by PGD with cross-entropy criterion and translation-invariant form \cite{dong2019evading}, DeepFool attack \cite{moosavi2016deepfool}, and C\&W attack \cite{carlini2017towards}. For semantic segmentation, we adopt BIM, DeepFool, and C\&W for evaluation; for object detection, ``cls", ``loc", and ``cls+loc" (simultaneously conducting classification and localization attacks) are employed for testing. 

We experiment with the transferable attack to demonstrate our method's effect. In such evaluation, attackers cannot utilize the exact gradient information of the target model. Instead, they usually obtain the gradient information from a substitute network, which is trained on the same dataset \cite{papernot2017practical,papernot2016transferability,liu2016delving} with different model structures. Thus, in evaluation of the classification task, we use the perturbations computed from ResNet50 to attack the defense framework trained with WideResNet. 
%For the semantic segmentation task, the adversarial samples obtained from DeepLabv3 are adopted to achieve attack for PSPNet. 
For semantic segmentation, the adversarial samples obtained from DeepLabv3 are adopted to achieve attack for PSPNet. 
As for object detection, attacks for SSD are implemented by employing the adversarial perturbations generated from RFBNet.

\vspace{0.05in}
\noindent\textbf{Attack parameters.}
Following \cite{zhang2019theoretically,xu2020dynamic,zhang2019towards}, for the classification task, we adopt PGD with cross-entropy criterion ($\epsilon=0.031 \times 255$, $\alpha=0.0075\times 255$, $n=8$), DeepFool attack ($\epsilon=0.031 \times 255$), C\&W attack ($\epsilon=0.031 \times 255$, the step size is $0.0075\times 255$).
For semantic segmentation, we utilize BIM ($\epsilon=0.03 \times 255$, $\alpha=0.01\times 255$, $n=4$), DeepFool attack ($\epsilon=0.03 \times 255$), C\&W attack ($\epsilon=0.03 \times 255$, the step size is $0.01\times 255$) for evaluation.
%Moreover, the attacks for classification loss and location loss are employed within the object detection task, and the perturbation range is 8 for pixel values within $[0, 255]$.
Moreover, the perturbation range is set as 8 for the object detection task.
Note that we adopt different attack types and parameters for training and evaluation, to verify the generalization ability of the trained DGNs towards unseen attacks.
%Attacks are with untargeted form and their details are given in the supplementary file.

\vspace{0.05in}
\noindent\textbf{Metrics.}
Classification accuracy, mean of class-wise intersection over union (miou) and mean average precision (mAP), are adopted as the quality indicators for image classification, semantic segmentation, and object detection, respectively.

\subsection{Comparison with Existing Methods}

%%\vspace{-0.1in}
\noindent\textbf{Baselines.} We choose approaches of input transformation and adversarial training for comparison.
Most existing adversarial training work focuses on image classification, and we adopt two recent methods,  \cite{zhang2019theoretically} and \cite{shafahi2019adversarial}, for comparison. For the semantic segmentation task, \cite{xu2020dynamic} first conducted comprehensive exploration on the impact of adversarial training for semantic segmentation. We employ two strategies proposed by \cite{xu2020dynamic} (SAT and DDC-AT).
For object detection, \cite{zhang2019towards} proposed four variants of adversarial training.
All these adversarial training approaches are trained with their original configuration and ``finetune" means finetuning pre-trained models (without defense) via adversarial training.
For input transformation strategies, we use six representative methods, including \cite{naseer2020self,liao2018defense,shen2017ape,li2020enhancing,samangouei2018defense,mustafa2019image}.
We adopt their original generator structures, and re-train them with the same epochs, batch size and target models as ours.

\begin{table*}[t]
	\centering
	\caption{The evaluation with various attacks on CIFAR10, CIFAR100 and Tiny-ImageNet.
		%, and we report the value of accuracy (\%). 
		``WideResNet $\rightarrow$ ResNet50": attacking WideResNet while generating adversarial samples from ResNet50 (except the decision-based attack that does not utilizing the transferability of adversarial samples).} 
	\vspace{-0.1in}
	\large
	\label{tab:box1}
	\resizebox{1\linewidth}{!}{
		\begin{tabular}{l|cccccccccccc}
			\toprule[1pt]
			\multirow{2}{2.8cm}{WideResNet $\rightarrow$ ResNet50}&\multicolumn{12}{c}{CIFAR10 (Accuracy \%)}\\
			\cline{2-13}
			& SSP \cite{naseer2020self} & FDA \cite{ganeshan2019fda}& TAP \cite{zhou2018transferable}&MI \cite{dong2018boosting} & Auto \cite{croce2020reliable}&Square \cite{andriushchenko2020square}  &Cu\&Wh \cite{shi2019curls} & DIM \cite{xie2019improving} &PPDA \cite{li2020projection} & GeoDA \cite{rahmati2020geoda}& SF \cite{chen2020boosting}  & DR \cite{lu2020enhancing}\\
			\hline
			{Denoise \cite{liao2018defense}} &60.2&65.6&67.1&71.4 &61.4&59.6 &58.4 &64.1 &58.2&59.5&56.2&56.9 \\
			{NRP \cite{naseer2020self}}  &62.0&66.2&69.2&71.9 &62.3&61.1&60.7&65.3 &59.7&60.3&57.7&58.4\\
			Ours &\textbf{64.1}&\textbf{68.0}&\textbf{70.1}&\textbf{73.2}&\textbf{65.2}&\textbf{63.3}&\textbf{62.5} &\textbf{67.4} &\textbf{63.4}&\textbf{64.7} &\textbf{61.3}&\textbf{63.9}\\
			
			\hline
			\multirow{2}{2.8cm}{WideResNet $\rightarrow$ ResNet50}&\multicolumn{12}{c}{CIFAR100 (Accuracy \%)}\\
			\cline{2-13}
			& SSP \cite{naseer2020self} & FDA \cite{ganeshan2019fda}& TAP \cite{zhou2018transferable}&MI \cite{dong2018boosting} & Auto \cite{croce2020reliable}&Square \cite{andriushchenko2020square}  &Cu\&Wh \cite{shi2019curls} & DIM \cite{xie2019improving} &PPDA \cite{li2020projection} & GeoDA \cite{rahmati2020geoda}& SF \cite{chen2020boosting}  & DR \cite{lu2020enhancing}\\
			\hline
			{Denoise \cite{liao2018defense}} &40.8&46.3&46.7&48.3 &43.2&40.6 &38.8 &40.5 &35.1&36.9&33.7&36.1 \\
			{NRP \cite{naseer2020self}}  &41.7&45.8&45.9&46.3&42.1&39.1 &40.2 & 42.3&38.8&40.4 &36.9&39.2\\
			Ours&\textbf{44.9}&\textbf{47.6}&\textbf{48.1}&\textbf{50.3}&\textbf{45.5}&\textbf{41.8}&\textbf{42.2}&\textbf{46.7} &\textbf{42.3}&\textbf{43.5}&\textbf{40.2}&\textbf{43.0}\\
			
			\hline
			\multirow{2}{2.8cm}{WideResNet $\rightarrow$ ResNet50}&\multicolumn{12}{c}{Tiny-ImageNet (Accuracy \%)}  \\
			\cline{2-13}
			& SSP \cite{naseer2020self} & FDA \cite{ganeshan2019fda}& TAP \cite{zhou2018transferable}&MI \cite{dong2018boosting} & Auto \cite{croce2020reliable}&Square \cite{andriushchenko2020square}  &Cu\&Wh \cite{shi2019curls} & DIM \cite{xie2019improving} &PPDA \cite{li2020projection} & GeoDA \cite{rahmati2020geoda}& SF \cite{chen2020boosting}  & DR \cite{lu2020enhancing}\\
			\hline
			{Denoise \cite{liao2018defense}} &34.5&\textbf{40.9}&38.5&41.2 &34.2&31.8 &31.7& 36.2&32.9&35.0&31.0&34.7 \\
			{NRP \cite{naseer2020self}}  & 35.5&39.1&37.2&40.6 &36.7&30.5 &32.4& 37.6&33.4&34.8 &31.3&35.1 \\
			Ours &\textbf{37.3}&40.6&\textbf{41.8}&\textbf{44.5} &\textbf{38.8}&\textbf{33.1} &\textbf{34.2}&\textbf{40.2} &\textbf{34.9}&\textbf{36.6}&\textbf{32.8}&\textbf{35.2} \\
			\bottomrule[1pt]
	\end{tabular}}
	\vspace{-0.1in}
\end{table*}

\begin{table}[t]
	\centering
	\caption{Results of statistically significant calculation for the comparison in Table \ref{tab:black-box0} and \ref{tab:black-box1}. We report the p-value.}
	\vspace{-0.1in}
	\Huge
	\label{tab:sign}
	\resizebox{1.0\linewidth}{!}{
		\begin{tabular}{l|p{3cm}<{\centering}p{3cm}<{\centering}p{3cm}<{\centering}|p{3cm}<{\centering}p{3cm}<{\centering}p{3cm}<{\centering}}
			\hline
			&\multicolumn{3}{c|}{CIFAR10} &\multicolumn{3}{c}{CIFAR100} \\
			\cline{2-7}
			& {PGD} & {DeepFool}& {C\&W} & {PGD} & {DeepFool}& {C\&W}  \\
			\hline
			Ours vs Denoise &2.7e-17&3.2e-8&5.6e-7&5.8e-17&4.3e-16&3.6e-7\\
			Ours vs NRP &1.6e-13&2.6e-7&7.4e-10&7.5e-9&6.8e-7&9.2e-12\\
			\hline
			&\multicolumn{3}{c|}{Tiny-ImageNet} &\multicolumn{3}{c}{Cityscapes} \\
			\cline{2-7}
			& {PGD} & {DeepFool}& {C\&W} & {BIM} & {DeepFool}& {C\&W}  \\
			\hline
			Ours vs Denoise &5.5e-7&8.2e-8&1.7e-7&7.2e-11&8.4e-10&2.9e-8\\
			Ours vs NRP &6.7e-8&2.4e-11&5.0e-10&7.3e-8&3.3e-10&9.1e-9\\
			\hline
			&\multicolumn{3}{c|}{VOC2012} &\multicolumn{3}{c}{VOC07+12} \\
			\cline{2-7}
			& {BIM} & {DeepFool}& {C\&W}& {cls} & {loc} & {cls+loc}  \\
			\hline
			Ours vs Denoise &2.6e-11&8.3e-7&9.4e-7&1.1e-8&9.8e-10&6.9e-7\\
			Ours vs NRP &4.4e-9&5.7e-7&1.8e-10&4.8e-9&7.1e-7&8.8e-10\\
			\hline
	\end{tabular}}
\vspace{-0.1in}
\end{table}

\begin{table*}[t]
	\centering
	%\caption{Quantitative comparison on the classification tasks in terms of the quality enhancement.  ``WideResNet $\rightarrow$ ResNet50": attacking WideResNet while generating adversarial samples from ResNet50.} 
	\caption{Comparison on classification, semantic segmentation, and object detection tasks in terms of quality enhancement.} 
		\vspace{-0.1in}
	\Huge
	\label{tab:black-box0-qa}
	\resizebox{1.0\linewidth}{!}{
		\begin{tabular}{l|p{3.5cm}<{\centering}p{3.5cm}<{\centering}p{3.5cm}<{\centering}p{3.5cm}<{\centering}|p{3.5cm}<{\centering}p{3.5cm}<{\centering}p{3.5cm}<{\centering}p{3.5cm}<{\centering}|p{3.5cm}<{\centering}p{3.5cm}<{\centering}p{3.5cm}<{\centering}p{3.5cm}<{\centering}}
			\toprule[1pt]
			\multirow{2}{5cm}{WideResNet $\rightarrow$ ResNet50}&\multicolumn{4}{c|}{CIFAR10 (dB)} &\multicolumn{4}{c|}{CIFAR100 (dB)} &\multicolumn{4}{c}{Tiny-ImageNet (dB)}  \\
			\cline{2-13}
			&clean& {PGD} & {DeepFool}& {C\&W} &clean& {PGD} & {DeepFool}& {C\&W} & clean&{PGD} & {DeepFool}& {C\&W}   \\
			\hline
			Defense \cite{samangouei2018defense} &23.70& 23.54&23.50 &23.64&20.86 &20.54&20.72&20.79&23.63&23.05&23.34&23.40 \\
			SR \cite{mustafa2019image}&24.58&24.05 &24.11 &24.26& 21.20&21.08&21.14&21.18&25.27&25.04&25.09&25.21 \\
			FPD \cite{li2020enhancing}&24.64& 24.16& 24.24&24.45&22.08 &21.85&21.91&21.93&26.14&25.67&26.06&26.09 \\
			APE \cite{shen2017ape} &27.52& 23.77&26.82 &26.91& \textbf{25.65}&21.46&23.88&24.15&28.45&25.30&28.01&28.15 \\
			Denoise \cite{liao2018defense}&27.43& 25.24&26.85 &26.98&24.87 &22.21&24.04&23.96&28.14&26.11&27.67&27.87 \\
			NRP \cite{naseer2020self} &\textbf{27.81}&25.03 &27.03 &27.07&25.32 &22.03&24.13&24.20&28.01&26.32&27.88&27.92 \\
			Ours &27.69&\textbf{25.52} &\textbf{27.17} &\textbf{27.26}& 25.09&\textbf{22.67}&\textbf{24.55}&\textbf{24.87}&\textbf{28.52} &\textbf{26.74}&\textbf{28.25}&\textbf{28.36} \\
			%\bottomrule[1pt]
			\hline
						PSPNet $\rightarrow$ DeepLabv3&\multicolumn{4}{c|}{Cityscapes (dB)} &\multicolumn{4}{c|}{VOC2012 (dB)} &\multicolumn{4}{c}{VOC07+12 (dB)}  \\
			\cline{2-13}
			SSD $\rightarrow$ RFBNet& clean& {BIM} & {DeepFool}& {C\&W}  &clean& {BIM} & {DeepFool}& {C\&W} &clean& {cls} & {loc} & {cls+loc}   \\
			\hline
			Defense \cite{samangouei2018defense}&21.06&20.52& 20.36&20.92 &20.83&20.08 &20.34&20.50&23.21&22.88&23.03&22.84 \\
			{SR \cite{mustafa2019image}} &28.14&26.16& 26.05& 26.13&29.14&26.06&26.31&28.02&24.46&23.34&23.52&23.19 \\
			FPD \cite{li2020enhancing}&27.70&26.00&26.19 &26.69 &27.38& 26.04&26.27&26.53&25.87&25.72&25.01&25.33 \\
			APE \cite{shen2017ape}&27.15&24.87&24.21& 25.38 &\textbf{30.06}&22.50 &25.80&26.15&\textbf{26.23}&25.06&25.07&24.80 \\
			Denoise \cite{liao2018defense}&28.93&27.19&27.04 &28.84 &29.02&26.71 &26.55&28.06&26.15&24.21&23.86&24.08 \\
			NRP \cite{naseer2020self}  &29.51&27.04&26.02&29.41 &29.07&26.85 &26.03&28.11&25.90&25.27&24.63&25.02 \\
			Ours &\textbf{30.02}&\textbf{27.45}& \textbf{28.93}& \textbf{29.66}&29.25&\textbf{27.13} &\textbf{28.22}&\textbf{28.64}&25.94&\textbf{25.81}&\textbf{25.08}&\textbf{25.37} \\
			\bottomrule[1pt]
	\end{tabular}}
%	\vspace{-0.1in}
\end{table*}

\vspace{0.05in}
\noindent\textbf{Quantitative results.} The results in the classification task are summarized in Table \ref{tab:black-box0}.
Although a few approaches yield comparable effect on clean samples, our full setting results in the highest robustness on adversarial samples, proving the effectiveness of our method.
Further, as exhibited in Table \ref{tab:black-box1}, in semantic segmentation and object detection, our results outperform all other competing methods on adversarial samples. 
Only a fraction of approaches work decently on clean samples. Most lack the similar level of robustness on adversarial samples (e.g., APE). 

We also conduct the comparison between our approach and current methods, with the target model structure of ResNet50 \cite{he2016deep}, DeepLabv3 \cite{chen2017rethinking} and RFBNet \cite{liu2018receptive}. The results are summarized in Table \ref{tab:black-box111} and \ref{tab:black-box444}, and these results all support the superiority of our approach over current methods.

\vspace{0.05in}
\noindent\textbf{Evaluation with various attacks on classification tasks.}
Furthermore, there are other state-of-the-art black-box attack methods that are designed for the classification task. 
We include state-of-the-art attacks for evaluation, including SSP attack \cite{naseer2020self}, MI-FGSM \cite{dong2018boosting}, FDA \cite{ganeshan2019fda}, TAP attack \cite{zhou2018transferable}, DIM attack \cite{xie2019improving}, Square attack \cite{andriushchenko2020square}, AutoAttack \cite{croce2020reliable}, Curls\&Whey (Cu\&Wh) attack \cite{shi2019curls}, PPDA attack \cite{li2020projection}, GeoDA attack \cite{rahmati2020geoda}, SF attack \cite{chen2020boosting} and Dispersion reduction (DR) attack \cite{lu2020enhancing}.
They are all implemented as the untargeted attacks.
The SSP attack is implemented with attack iteration number as 100, perturbation range as $\epsilon=0.031 \times 255$, step size as $\alpha=0.0075\times 255$, and the attack is completed by using the VGG16 layers.
The MI-FGSM attack is implemented with attack iteration number as 100, perturbation range as $\epsilon=0.031 \times 255$, step size as $\alpha=0.0075\times 255$, and decay factor as 1.0. 
The FDA attack, TAP attack and DIM attack are implemented with attack iteration number as 100, perturbation range as $\epsilon=0.031 \times 255$, and step size as $\alpha=0.0075\times 255$. Especially, the DIM attack is implemented with translation invariant attack \cite{dong2019evading} with decay factor as 1.0, the probability for stochastic transformation function as 0.5.
The Square attack is completed with $\epsilon=0.031 \times 255$ and attack iteration number as 10000. For the Square attack, loss function is adopted as the cross-entropy loss and margin loss, and the probability of changing a coordinate is set as 0.3.
The AutoAttack is implemented with the PGD attack using the cross-entropy loss, and attack iteration number is set as 1000, number of restart is set as 5, $\rho$ is set as 0.75, and $\epsilon=0.031 \times 255$.
The AutoAttack utilizes the transferability of adversarial samples.
Curls\&Whey attack is implemented by setting attack iteration number as 100, the variance of gaussian noise as 2, the step size for attack as $0.0075 \times 255$, the binary search step as 12, and perturbation range as $\epsilon=0.031 \times 255$. 
The PPDA attack is completed with maximal iteration number as 4000, the size for the reduction of dimension is set as 1500, the amplitude of change at each step is set as $0.0075 \times 255$, and the perturbation range is $\epsilon=0.031 \times 255$
For GeoDA attack, the maximal iteration number is 4000, $\lambda=0.6$, the dimension of the subspace is 75, and the perturbation range is $\epsilon=0.031 \times 255$
For SF attack, the maximal iteration number is 20000 and the perturbation range is $\epsilon=0.031 \times 255$ and the the dimensionality reduction rate is 2.0.
For DR attack, the attack iteration number is 100, perturbation range is $\epsilon=0.031 \times 255$, step size as $\alpha=0.0075\times 255$.
The results of our method and two strong baselines (Denoise \cite{liao2018defense} and NRP \cite{naseer2020self}) are reported in Table \ref{tab:box1}. 
These results demonstrate that our approach can keep robustness towards various types of attacks and outperforms these two strong baselines.

\begin{figure*}[t]
	\centering
	\newcommand\widthpose{0.39}
	\newcommand\heightpos{0.25}
	\newcommand\heightpose{0.213}
	\newcommand\heightposee{0.253}
	\newcommand\heightposeee{0.25}
	\resizebox{1.0\linewidth}{!}{
		\begin{tabular}{@{\hspace{0.0mm}}c@{\hspace{1.0mm}}c@{\hspace{1.0mm}}c@{\hspace{1.0mm}}c@{\hspace{1.0mm}}c@{\hspace{1.0mm}}c@{\hspace{0.0mm}}}

			\specialrule{0em}{3pt}{3pt}
			\includegraphics[align=c,width=\widthpose\textwidth]{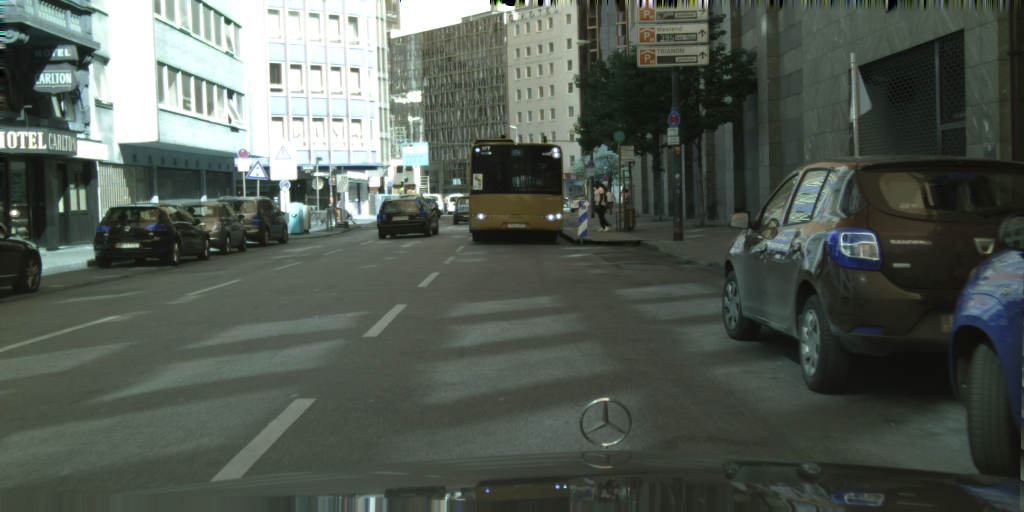} &
			\includegraphics[align=c,width=\widthpose\textwidth]{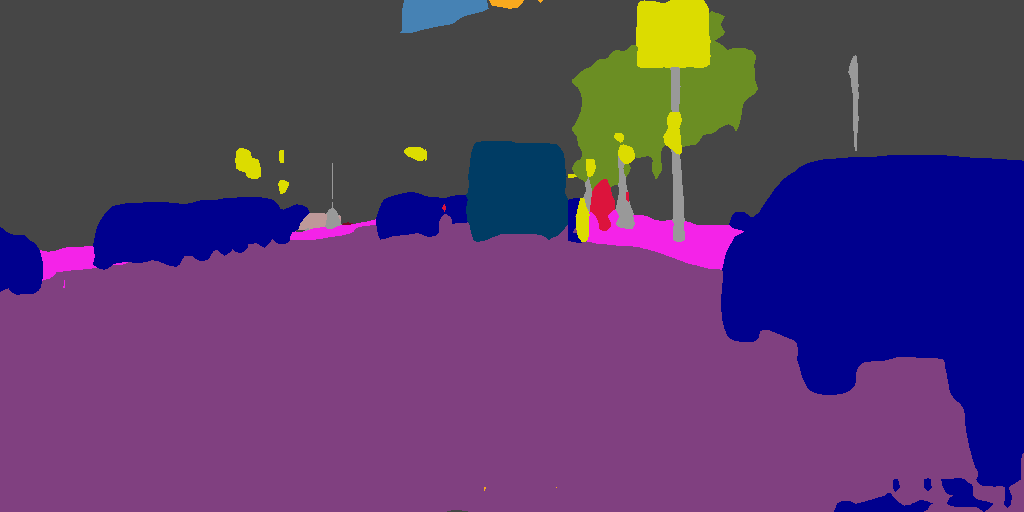} &
			\includegraphics[align=c,width=\widthpose\textwidth]{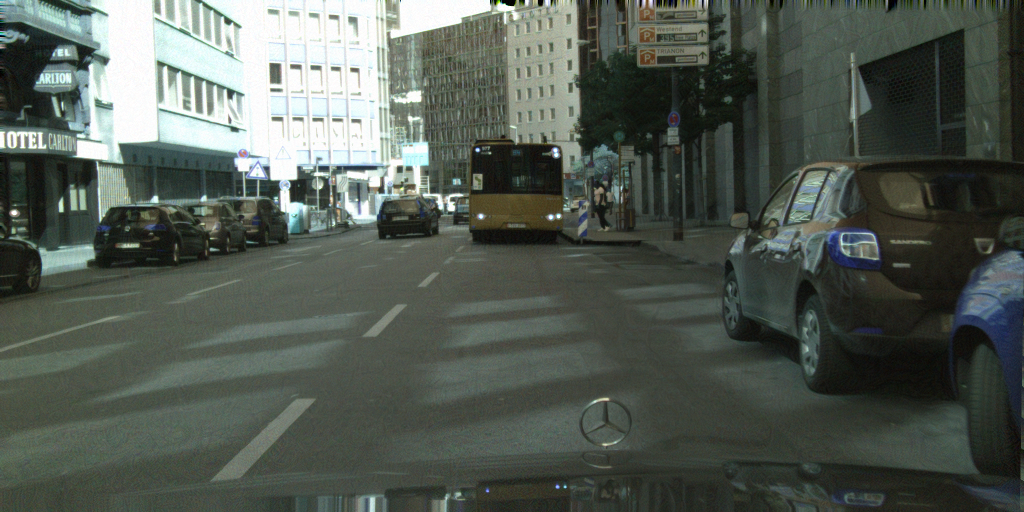} &
			\includegraphics[align=c,width=\widthpose\textwidth]{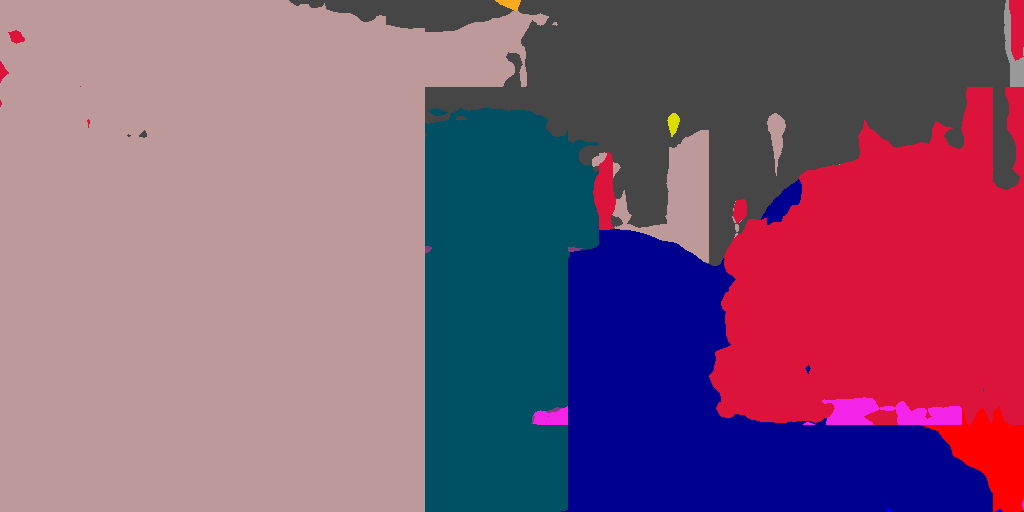} &
			\includegraphics[align=c,width=\widthpose\textwidth]{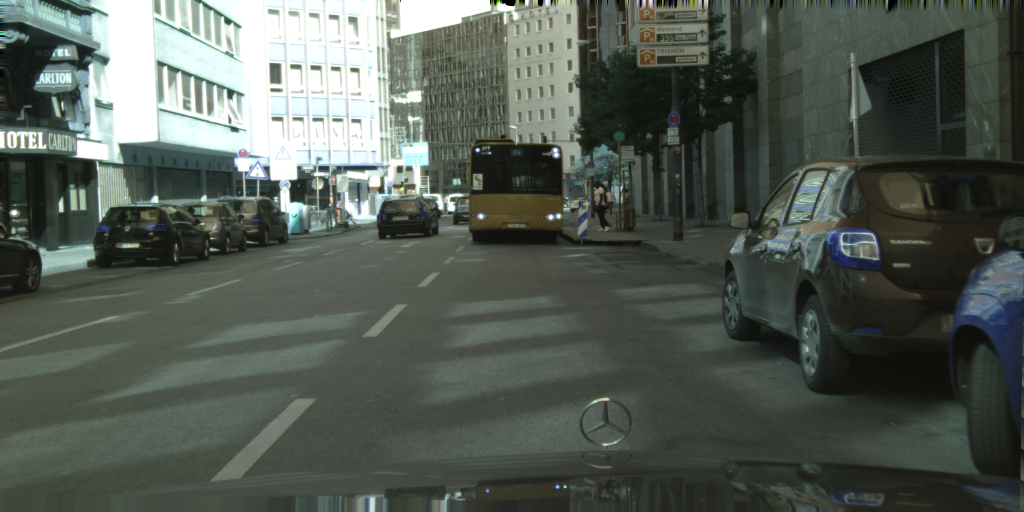}&
			\includegraphics[align=c,width=\widthpose\textwidth]{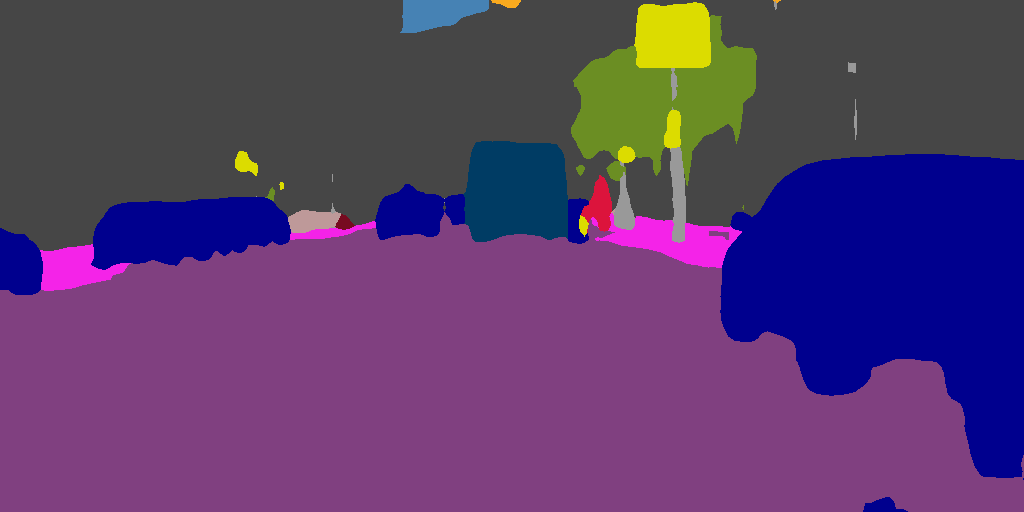} \\
			
			\specialrule{0em}{3pt}{3pt}
			\includegraphics[width=\widthpose\textwidth]{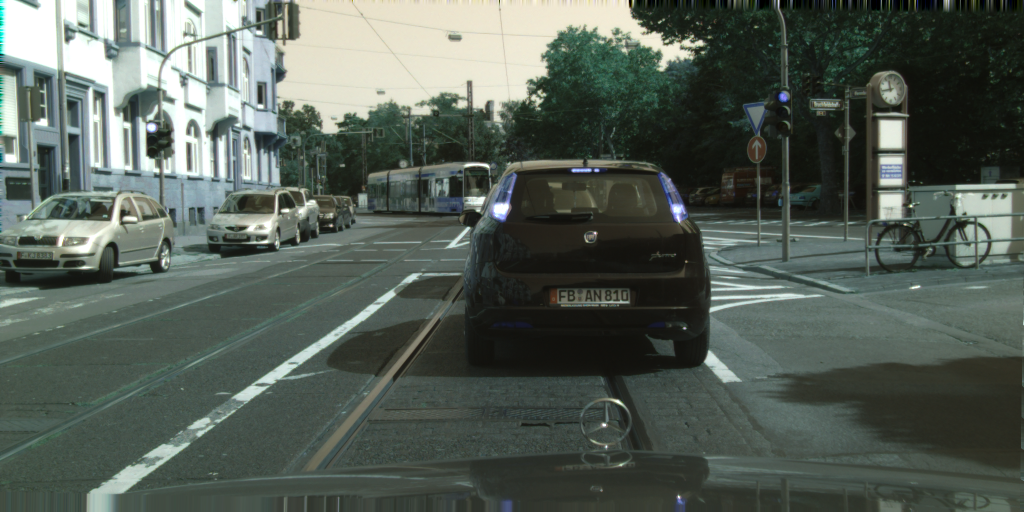} &
			\includegraphics[width=\widthpose\textwidth]{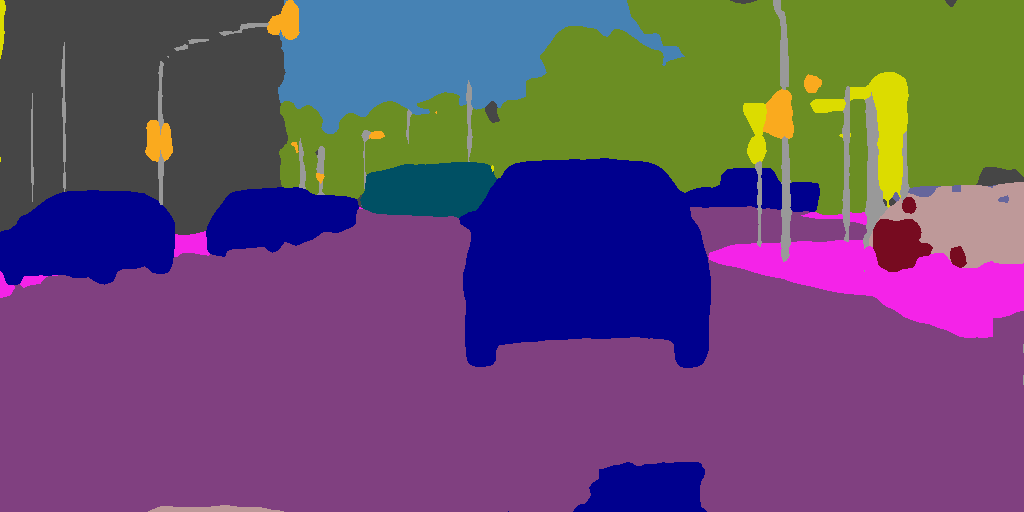} &
			\includegraphics[width=\widthpose\textwidth]{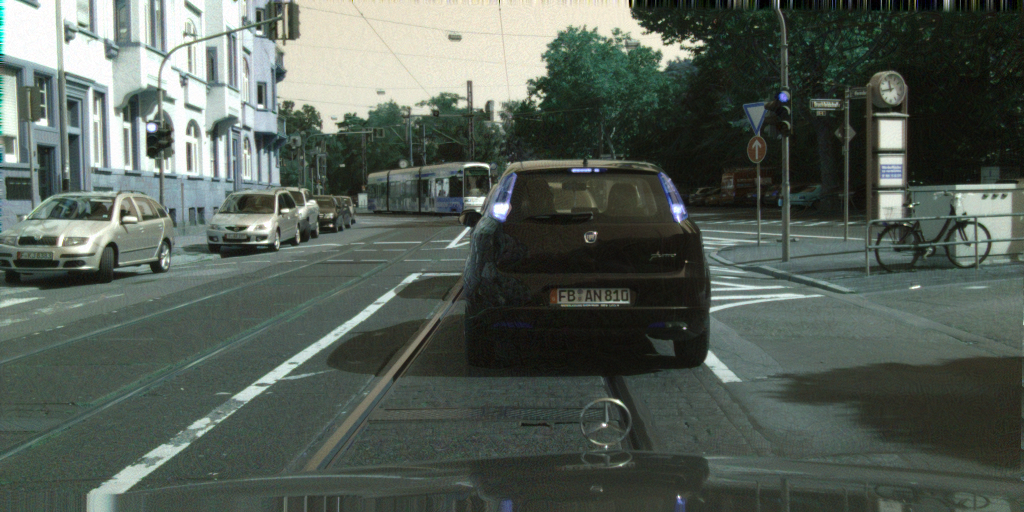} &
			\includegraphics[width=\widthpose\textwidth]{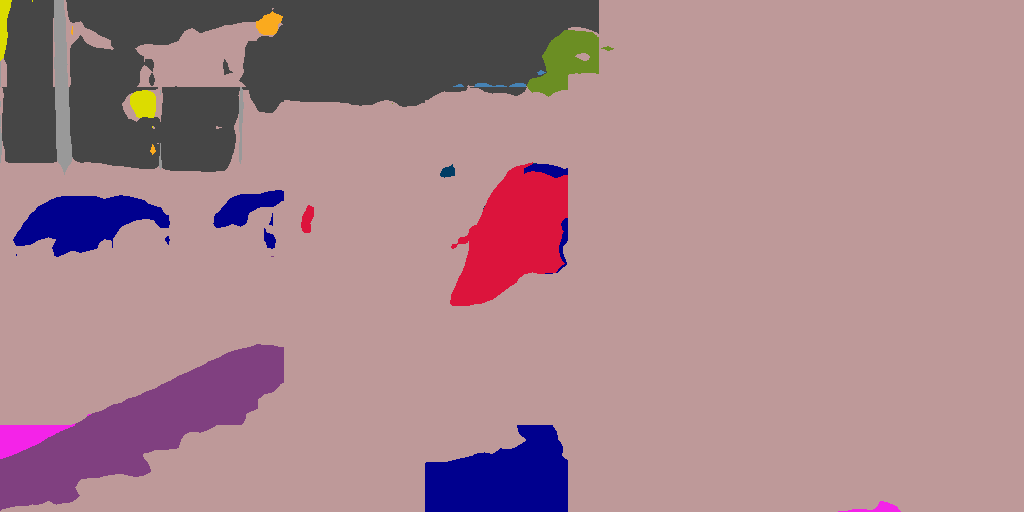} &
			\includegraphics[width=\widthpose\textwidth]{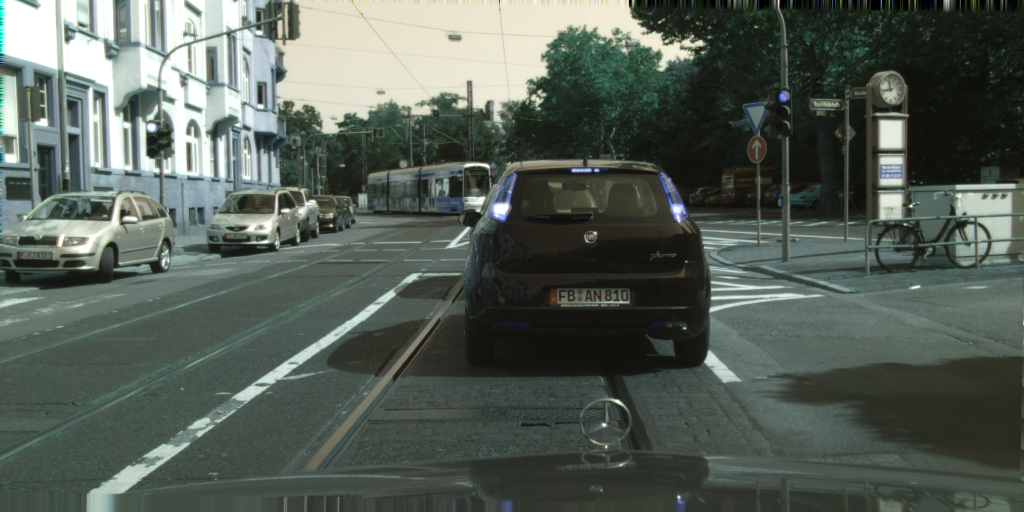}&
			\includegraphics[width=\widthpose\textwidth]{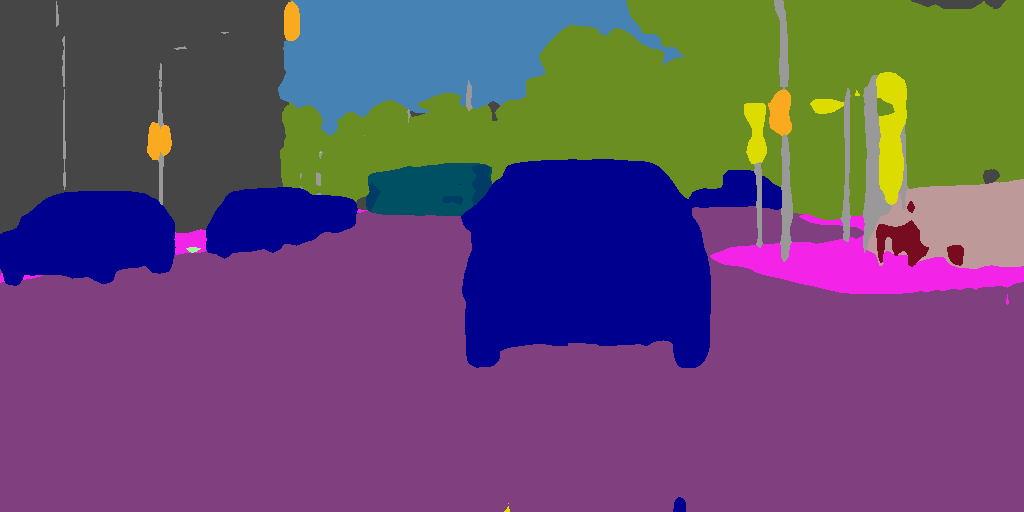} \\
			
			\specialrule{0em}{3pt}{3pt}
			\includegraphics[width=\widthpose\textwidth, height=\heightpose\textwidth]{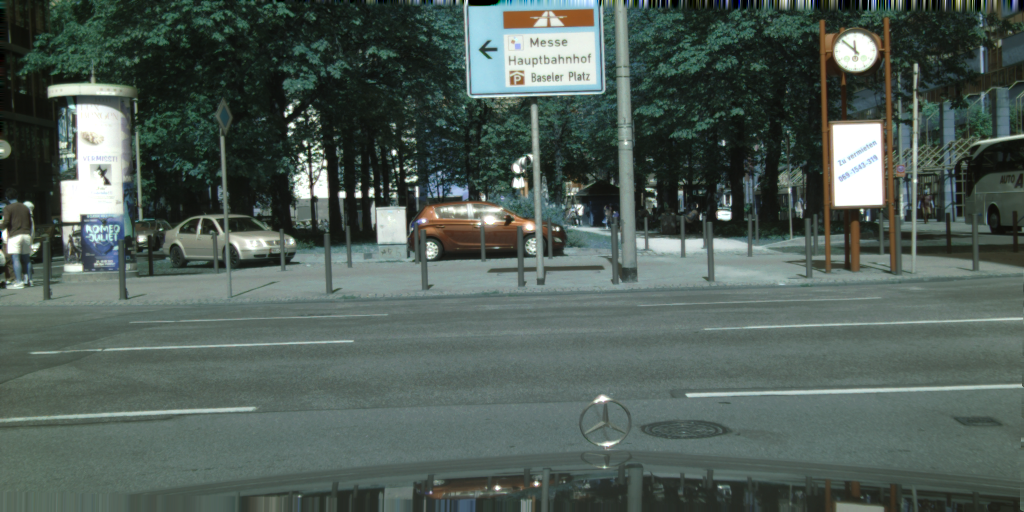} &
			\includegraphics[width=\widthpose\textwidth, height=\heightpose\textwidth]{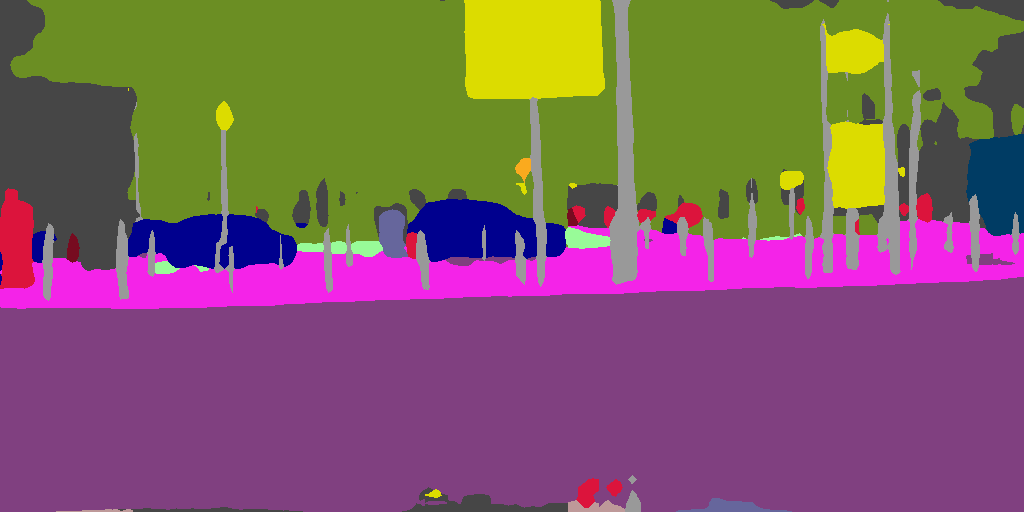} &
			\includegraphics[width=\widthpose\textwidth, height=\heightpose\textwidth]{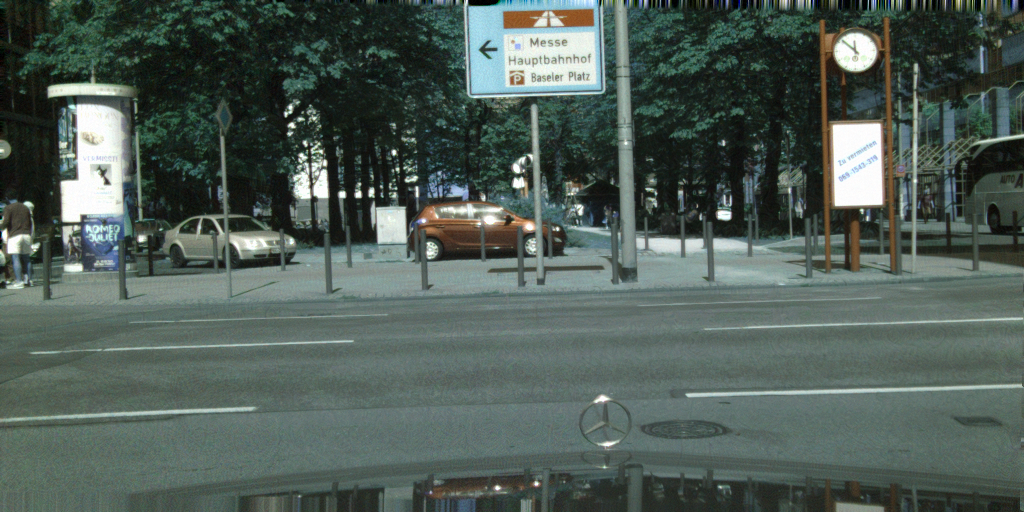} &
			\includegraphics[width=\widthpose\textwidth, height=\heightpose\textwidth]{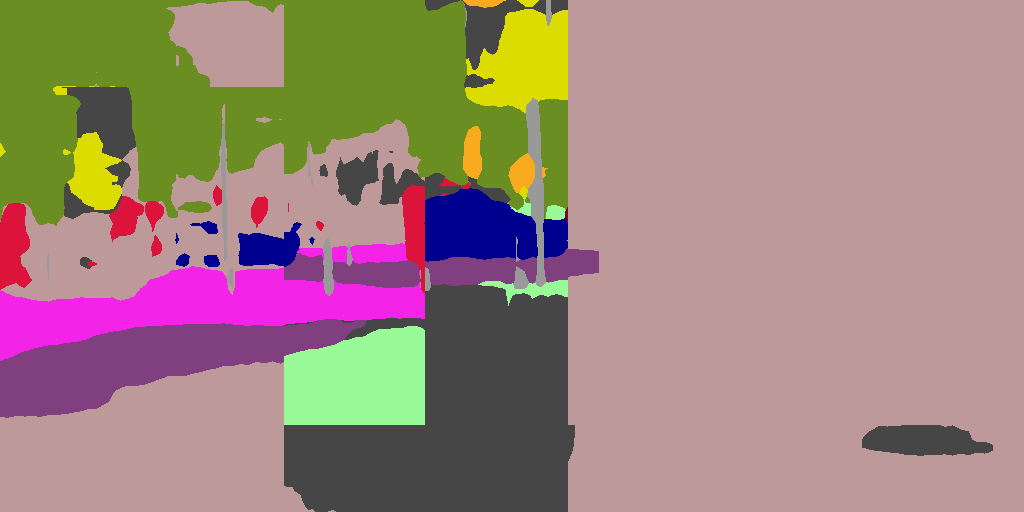} &
			\includegraphics[width=\widthpose\textwidth, height=\heightpose\textwidth]{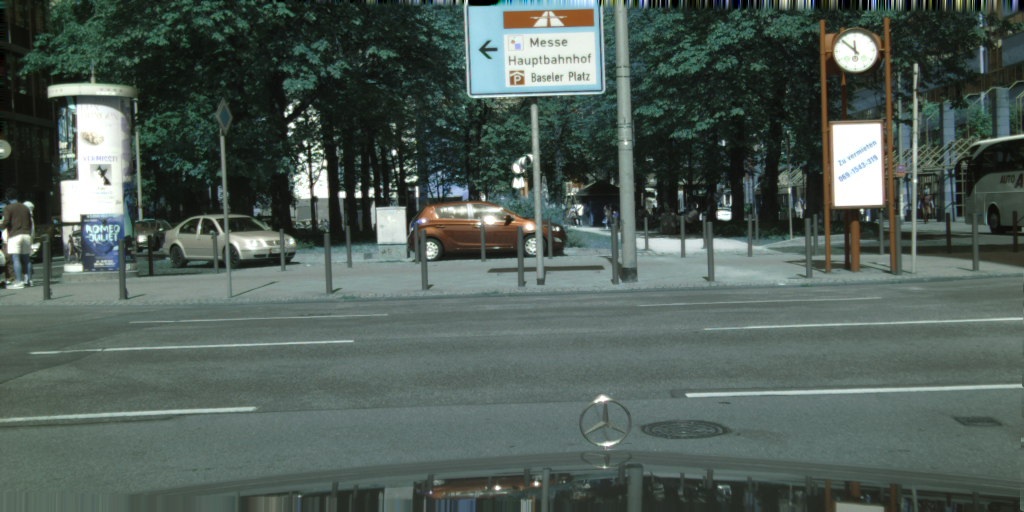}&
			\includegraphics[width=\widthpose\textwidth, height=\heightpose\textwidth]{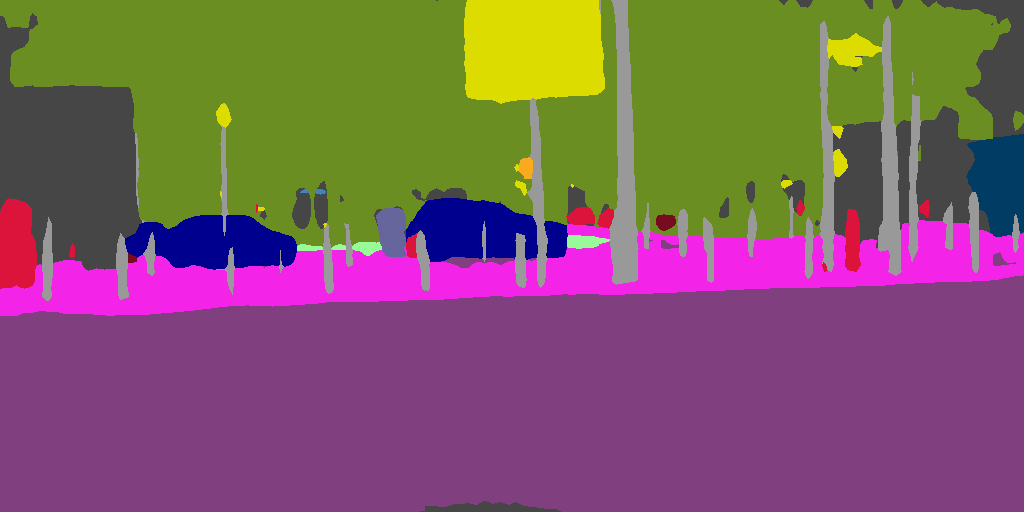} \\
			
			\specialrule{0em}{5pt}{5pt}
			\textbf{\Huge{$x^c$}}& {\Huge{result of $x^c$}}&\textbf{\Huge{$x^a$}}& {\Huge{result of $x^a$}}&\textbf{\Huge{$\widehat{x}^a$}}& {\Huge{result of $\widehat{x}^a$}}\\
	\end{tabular}}
	\caption{Visual illustration of results on clean samples $x^c$, adversarial samples $x^a$ (PGD attack) and processed adversarial samples $\widehat{x}^a$ with our trained $\mathcal{G}$, for the tasks of semantic segmentation on Cityscapes.}
	\vspace{-0.1in}
	\label{fig:visual2}
\end{figure*}

\begin{figure}[t]
	\centering
	\resizebox{1.0\linewidth}{!}{
		\begin{tabular}{@{\hspace{0.0mm}}c@{\hspace{8.0mm}}c@{\hspace{8.0mm}}c@{\hspace{8.0mm}}c@{\hspace{0.0mm}}}
			
			\multicolumn{4}{c}{t-SNE visualizations for classification task on CIFAR10  \cite{krizhevsky2009learning}}\\
			\includegraphics[width=0.18\linewidth]{images/classification/clean_clean2.jpg}&
			\includegraphics[width=0.18\linewidth]{images/classification/clean_adver2.jpg}&
			\includegraphics[width=0.18\linewidth]{images/classification/generate_clean2.jpg}&
			\includegraphics[width=0.18\linewidth]{images/classification/generate_adver2.jpg}\\
			&&&\\
			\multicolumn{4}{c}{t-SNE visualizations for segmentation task on Cityscapes \cite{cordts2016cityscapes}}\\
			\includegraphics[width=0.18\linewidth, height=0.18\linewidth]{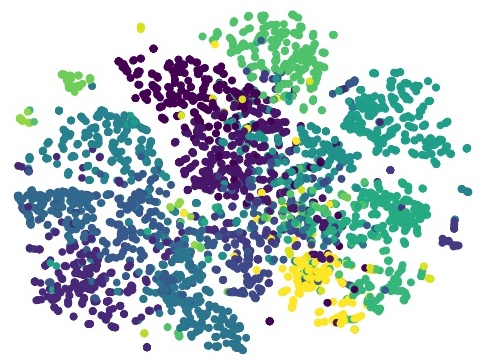}&
			\includegraphics[width=0.18\linewidth, height=0.18\linewidth]{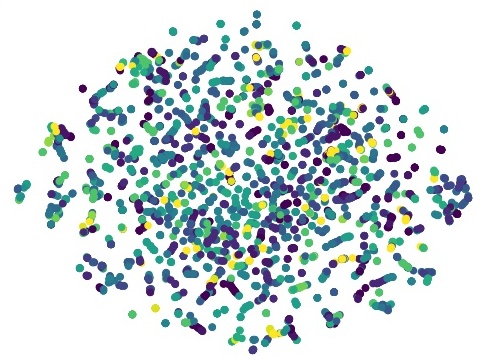}&
			\includegraphics[width=0.18\linewidth, height=0.18\linewidth]{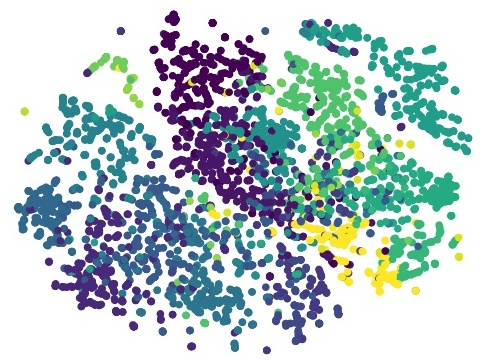}&
			\includegraphics[width=0.18\linewidth, height=0.18\linewidth]{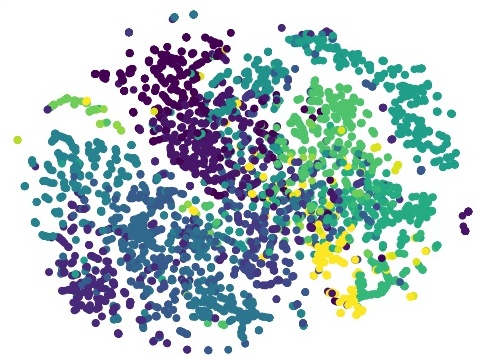}\\
			&&&\\
			\multicolumn{4}{c}{t-SNE visualizations for detection task on VOC07+12 \cite{everingham2010pascal}}\\
			\includegraphics[width=0.18\linewidth, height=0.18\linewidth]{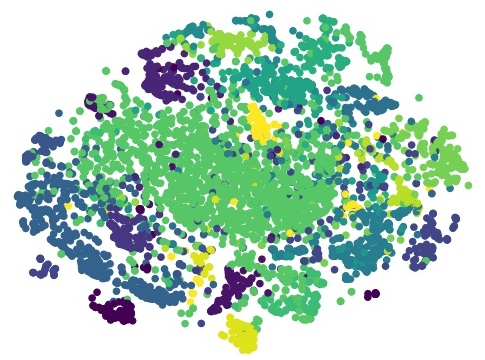}&
			\includegraphics[width=0.18\linewidth, height=0.18\linewidth]{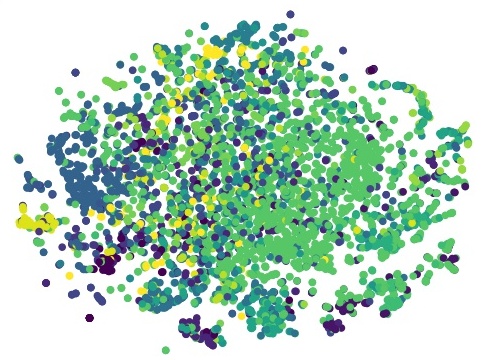}&
			\includegraphics[width=0.18\linewidth, height=0.18\linewidth]{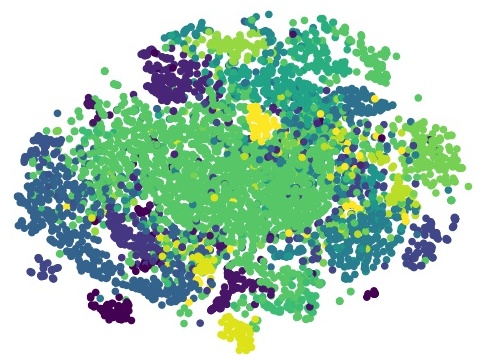}&
			\includegraphics[width=0.18\linewidth, height=0.18\linewidth]{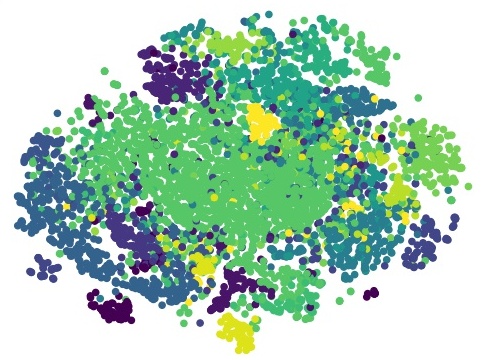}\\
			(a) $x^c$ in $\mathcal{O}$ & (b) $x^a$ in $\mathcal{O}$& (c) $\widehat{x}^c$ in $\mathcal{O}$  &(d) $\widehat{x}^a$ in $\mathcal{O}$ \\
	\end{tabular}}
\vspace{-0.1in}
	\caption{Target model $\mathcal{O}$ has ideal distributions for clean samples $x^c$ while disordered distributions for adversarial samples $x^a$. Our generator $\mathcal{G}$ can turn $x^c$/$x^a$ into $\widehat{x}^c$/$\widehat{x}^a$ with corrected distrubutions.}
	\vspace{-0.15in}
	\label{fig:short_tsne}
\end{figure}

\begin{figure}[t]
	\centering
	\newcommand\widthpose{0.25}
	\newcommand\heightpos{0.25}
	\newcommand\heightpose{0.213}
	\newcommand\heightposee{0.253}
	\newcommand\heightposeee{0.25}
	\resizebox{1.0\linewidth}{!}{
		\begin{tabular}{@{\hspace{0.0mm}}c@{\hspace{5.0mm}}c@{\hspace{5.0mm}}c@{\hspace{15.0mm}}c@{\hspace{15.0mm}}c@{\hspace{5.0mm}}c@{\hspace{0.0mm}}}
			\includegraphics[align=c,width=\widthpose\textwidth]{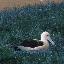}&
			\makecell[c]{\Huge{albatross, $\surd$} \\ \\ \Huge{score: 0.9998}}&
			\includegraphics[align=c,width=\widthpose\textwidth]{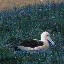}&
			\makecell[c]{\Huge{goose, $\times$} \\ \\ \Huge{score: 1.000}}&
			\includegraphics[align=c,width=\widthpose\textwidth]{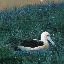}&
			\makecell[c]{\Huge{albatross, $\surd$} \\ \\ \Huge{score: 0.9887}}\\
			
			\specialrule{0em}{3pt}{3pt}
			\includegraphics[align=c,width=\widthpose\textwidth]{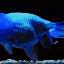}&
			\makecell[c]{\Huge{goldfish, $\surd$} \\ \\ \Huge{score: 0.9998}}&
			\includegraphics[align=c,width=\widthpose\textwidth]{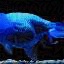}&
			\makecell[c]{\Huge{oboe, $\times$} \\ \\ \Huge{score: 1.000}}&
			\includegraphics[align=c,width=\widthpose\textwidth]{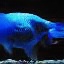}&
			\makecell[c]{\Huge{goldfish, $\surd$} \\ \\ \Huge{score: 0.9973}}\\
			
			%%%\specialrule{0em}{3pt}{3pt}
			%%%\includegraphics[align=c,width=\widthpose\textwidth]{images/visual/classification/n02106662_n02106662_0.9999_clean285.jpg}&
			%%%\makecell[c]{\Huge{german shepherd, $\surd$} \\ \\ \Huge{score: 0.9999}}&
			%%%\includegraphics[align=c,width=\widthpose\textwidth]{images/visual/classification/n02106662_n02085620_1.0000_adver285.jpg}&
			%%%\makecell[c]{\Huge{chihuahua, $\times$} \\ \\ \Huge{score: 1.000}}&
			%%%\includegraphics[align=c,width=\widthpose\textwidth]{images/visual/classification/n02106662_n02106662_0.9899_generated285.jpg}&
			%%%\makecell[c]{\Huge{german shepherd, $\surd$} \\ \\ \Huge{score: 0.9899}}\\
			
			\specialrule{0em}{3pt}{3pt}
			\textbf{\Huge{$x^c$}}& {\Huge{result of $x^c$}}&\textbf{\Huge{$x^a$}}& {\Huge{result of $x^a$}}&\textbf{\Huge{$\widehat{x}^a$}}& {\Huge{result of $\widehat{x}^a$}}\\
	\end{tabular}}
	\caption{Visual illustration of results on clean samples $x^c$, adversarial samples $x^a$ (PGD attack) and processed adversarial samples $\widehat{x}^a$ with our trained $\mathcal{G}$, for the tasks of classification on Tiny-ImageNet.}
	\vspace{-0.15in}
	\label{fig:visual}
\end{figure}

\begin{figure*}[t]
	\centering
	\newcommand\widthpose{0.39}
	\newcommand\heightpos{0.25}
	\newcommand\heightpose{0.213}
	\newcommand\heightposee{0.253}
	\newcommand\heightposeee{0.25}
	\resizebox{1.0\linewidth}{!}{
		\begin{tabular}{@{\hspace{0.0mm}}c@{\hspace{1.0mm}}c@{\hspace{1.0mm}}c@{\hspace{1.0mm}}c@{\hspace{1.0mm}}c@{\hspace{1.0mm}}c@{\hspace{0.0mm}}}
			
			\includegraphics[align=c,width=\widthpose\textwidth]{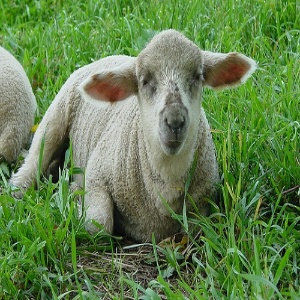} &
			\includegraphics[align=c,width=\widthpose\textwidth]{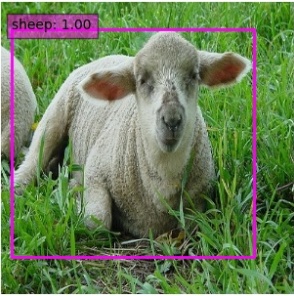} &
			\includegraphics[align=c,width=\widthpose\textwidth]{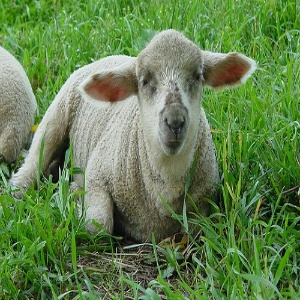} &
			\includegraphics[align=c,width=\widthpose\textwidth]{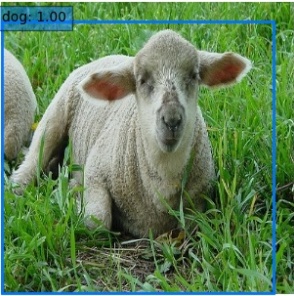} &
			\includegraphics[align=c,width=\widthpose\textwidth]{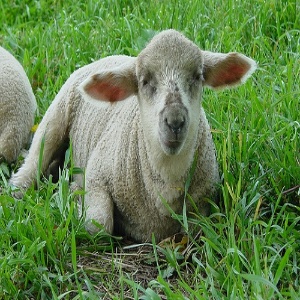}&
			\includegraphics[align=c,width=\widthpose\textwidth]{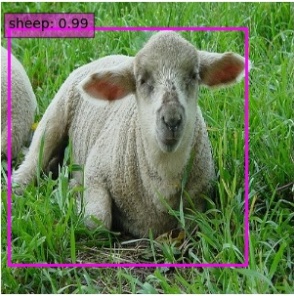} \\
			
			%%%\specialrule{0em}{5pt}{5pt}
			%%%\includegraphics[align=c,width=\widthpose\textwidth]{images/visual/detection/clean_47.jpg} &
			%%%\includegraphics[align=c,width=\widthpose\textwidth]{images/visual/detection/results_clean_47.jpg} &
			%%%\includegraphics[align=c,width=\widthpose\textwidth]{images/visual/detection/adver_47.jpg} &
			%%%\includegraphics[align=c,width=\widthpose\textwidth]{images/visual/detection/results_adver_47.jpg} &
			%%%\includegraphics[align=c,width=\widthpose\textwidth]{images/visual/detection/generated_47.jpg}&
			%%%	\includegraphics[align=c,width=\widthpose\textwidth]{images/visual/detection/results_generated_47.jpg} \\

			\specialrule{0em}{5pt}{5pt}
			\includegraphics[width=\widthpose\textwidth]{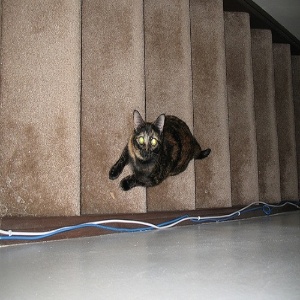} &
			\includegraphics[width=\widthpose\textwidth]{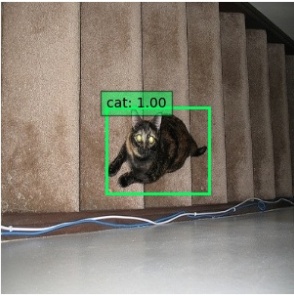} &
			\includegraphics[width=\widthpose\textwidth]{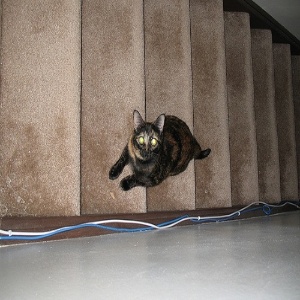} &
			\includegraphics[width=\widthpose\textwidth]{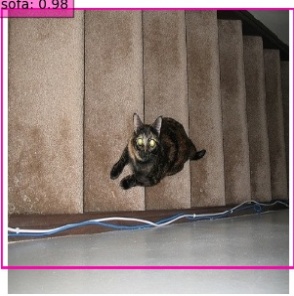} &
			\includegraphics[width=\widthpose\textwidth]{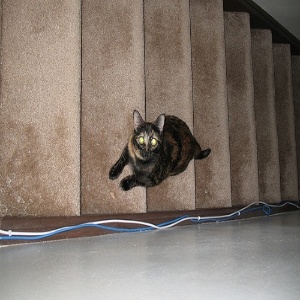}&
			\includegraphics[width=\widthpose\textwidth]{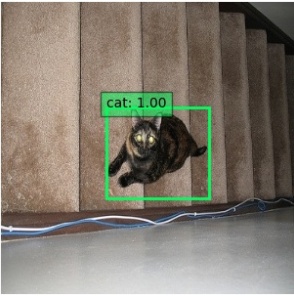} \\
			
			\specialrule{0em}{5pt}{5pt}
			\includegraphics[width=\widthpose\textwidth]{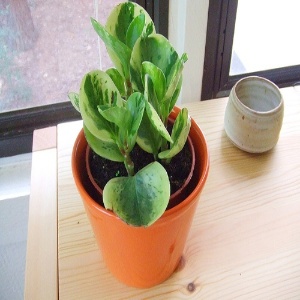} &
			\includegraphics[width=\widthpose\textwidth]{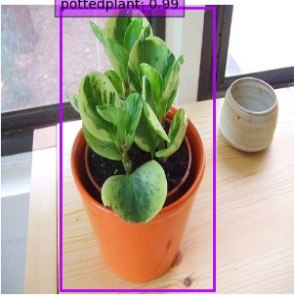} &
			\includegraphics[width=\widthpose\textwidth]{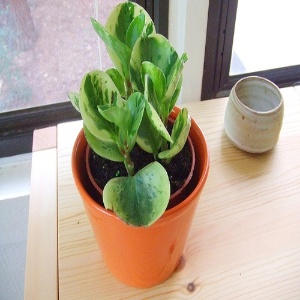} &
			\includegraphics[width=\widthpose\textwidth]{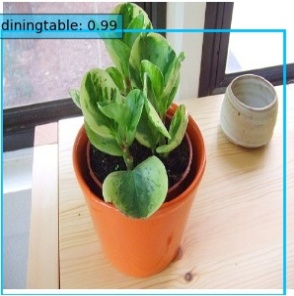} &
			\includegraphics[width=\widthpose\textwidth]{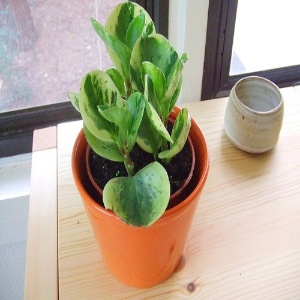}&
			\includegraphics[width=\widthpose\textwidth]{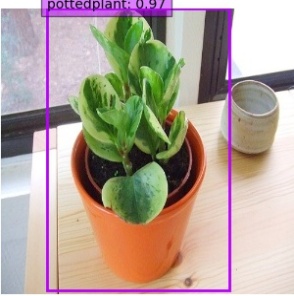} \\
			
			\specialrule{0em}{5pt}{5pt}
			\textbf{\Huge{$x^c$}}& {\Huge{result of $x^c$}}&\textbf{\Huge{$x^a$}}& {\Huge{result of $x^a$}}&\textbf{\Huge{$\widehat{x}^a$}}& {\Huge{result of $\widehat{x}^a$}}\\
	\end{tabular}}
	\caption{Visual illustration of results on clean samples $x^c$, adversarial samples $x^a$ (cls+loc attack) and processed adversarial samples $\widehat{x}^a$ with our trained $\mathcal{G}$, for the tasks of object detection on VOC07+12.}
	\label{fig:visual3}
\end{figure*}

\begin{table*}[t]
	\centering
	%\caption{Results of the ablation study. ``WideResNet $\rightarrow$ ResNet50": attacking WideResNet while generating adversarial samples from ResNet50; ``PSPNet $\rightarrow$ DeepLabv3": attacking PSPNet while obtaining adversarial samples with DeepLabv3; ``SSD $\rightarrow$ RFBNet": attacking SSD while acquiring adversarial samples from RFBNet.} 
	\caption{Results of the ablation study for effects of ``novel pixel-level constraints" and ``integrated distribution alignment".} 
	\vspace{-0.1in}
	\Huge
	\label{tab:black-box}
	\resizebox{1.0\linewidth}{!}{
		\begin{tabular}{l|p{3.5cm}<{\centering}p{3.5cm}<{\centering}p{3.5cm}<{\centering}p{3.5cm}<{\centering}|p{3.5cm}<{\centering}p{3.5cm}<{\centering}p{3.5cm}<{\centering}p{3.5cm}<{\centering}|p{3.5cm}<{\centering}p{3.5cm}<{\centering}p{3.5cm}<{\centering}p{3.5cm}<{\centering}}
			\toprule[1pt]
			\multirow{2}{5cm}{WideResNet $\rightarrow$ ResNet50}&\multicolumn{4}{c|}{CIFAR10 (Accuracy \%)} &\multicolumn{4}{c|}{CIFAR100 (Accuracy \%)} &\multicolumn{4}{c}{Tiny-ImageNet (Accuracy \%)}  \\
			\cline{2-13}
			& {clean} & {PGD} & {DeepFool}& {C\&W} & {clean} & {PGD} & {DeepFool}& {C\&W} & {clean} & {PGD} & {DeepFool}& {C\&W}   \\
			\hline
			No Defense & \textbf{95.1}&2.1&5.3&6.4&\textbf{78.1}&7.5&9.3&10.4&\textbf{64.5}&19.2&20.4 &21.2  \\
			{$\mathcal{L}_I$}  &90.4&66.4&85.3&86.1&67.9&46.4&66.6&67.5&63.4&34.3 &60.7 &61.7\\
			{$\mathcal{L}_{I+F(w/o \ c)}$}&90.7&78.2&89.0&89.5&68.1&53.4&67.0&67.8&63.2&46.1 &61.5 &62.4 \\
			{$\mathcal{L}_T$}&92.9&49.5&84.4&85.5&71.2&39.3&65.1&66.6&63.4&32.0 &58.3 &59.2  \\
			{$\mathcal{L}_{T+F(w/o \ c)}$}&92.5&74.1&87.6&88.3&70.3&50.4&65.9&66.8&63.2&44.4 &60.2 &61.1 \\
			Full&90.7&\textbf{81.4}&\textbf{90.4}&\textbf{90.6}&68.7&\textbf{54.0}&\textbf{67.5}&\textbf{68.4}&62.9&\textbf{46.4} &\textbf{62.2} &\textbf{63.2}\\
			\hline
			PSPNet $\rightarrow$ DeepLabv3&\multicolumn{4}{c|}{Cityscapes (mIoU \%)} &\multicolumn{4}{c|}{VOC2012 (mIoU \%)} &\multicolumn{4}{c}{VOC07+12 (mAP \%)}  \\
			\cline{2-13}
			SSD $\rightarrow$ RFBNet& {clean} & {BIM} & {DeepFool}& {C\&W} & {clean} & {BIM} & {DeepFool}& {C\&W} & {clean} & {cls} & {loc} & {cls+loc}    \\
			\hline
			No Defense &\textbf{73.5}&3.6&38.6&12.5& \textbf{76.4}&11.1 &46.1&16.6  &\textbf{72.5}&17.9 &14.5 &15.0 \\
			{$\mathcal{L}_I$}&61.5&50.8&52.4&56.1&73.1 &59.8 &63.0& 68.3 &63.4&59.4 &54.4 &58.2 \\
			{$\mathcal{L}_{I+F(w/o \ c)}$}&64.9&58.7&60.7&63.4& 70.3&62.9 &64.7&68.6  &58.7&60.7 &57.4 &60.1 \\
			{$\mathcal{L}_T$}&63.8&46.3&50.9&54.6& 74.1& 45.0&57.2&63.1  &65.1& 58.3&53.7 &56.3 \\
			{$\mathcal{L}_{T+F(w/o \ c)}$}&65.0&55.6&59.1&61.3&71.0 &61.3 &62.4&66.1  &60.0&59.6 &55.0 &58.2 \\
			Full&67.6&\textbf{59.5}&\textbf{62.0}&\textbf{64.7}&71.2 &\textbf{63.6} &\textbf{66.0}& \textbf{69.5} &60.5& \textbf{61.2}&\textbf{58.3} &\textbf{60.9} \\
			%%%%%%%%%%%%%%%%%%%%%%%%%%%%%%%%%%%%%%%
			\bottomrule[1pt]
			%\hline
	\end{tabular}}
\vspace{-0.1in}
%\vspace{0.05in}
\end{table*}

\vspace{0.05in}
\noindent\textbf{The statistically significant calculation.}
To further analyze the superiority of our method over baselines, we conduct a paired t-test for the results in Table \ref{tab:black-box0} and \ref{tab:black-box1} with the null hypothesis H0: ``the scores of two different methods do not have obvious difference." We use a significant level of 0.001, given that H0 is true, and we calculate the p-value as shown in Table \ref{tab:sign} by using the Microsoft Excel T-TEST function.
We find that all p-values under different attacks are smaller than 0.001. Hence, we can reject H0 and show that our result is different from others with a significant level of 0.001.

\vspace{0.05in}
\noindent\textbf{The evaluation for quality enhancement.}
For the input transformation strategy, we also need to compare the results of image quality enhancement, i.e., we can compute the PSNR value between the transformed adversarial samples $\widehat{x}^a$ and the clean samples $x^c$. And we list the comparison between our method and other input transformation strategies in Table~\ref{tab:black-box0-qa}. %and~\ref{tab:black-box1-qa}. 
Obviously, our method also has the superiority in terms of the image quality enhancement.

\vspace{0.05in}
\noindent\textbf{Distribution visualization results.} Furthermore, we provide the t-SNE visualizations for the classification task, the semantic segmentation task and object detection tasks as shown in Fig. \ref{fig:short_tsne}. The adversarial perturbations are obtained with the PGD attack, BIM attack and “cls+loc” attack.

\vspace{0.05in}
\noindent\textbf{Qualitative results.} We provide visual illustrations to demonstrate the defense effects of our framework.
% on different tasks. %The visual illustrations for image classification are shown in Fig. \ref{fig:visual}, the visual cases for semantic segmentation are displayed in Fig. \ref{fig:visual2} , and the visual examples for object detection are exhibited in Fig. \ref{fig:visual3}. 
The visual cases for semantic segmentation are displayed in Fig. \ref{fig:visual2} , for image classification are shown in Fig. \ref{fig:visual}, and for object detection are exhibited in Fig. \ref{fig:visual3}.
Compared with the situations without our defense, our trained $\mathcal{G}$ leads to satisfying defense on adversarial samples in all three tasks.

\begin{table*}[t]
	\centering
	%\caption{\xgxu{Results of the ablation study for loss functions. ``WideResNet $\rightarrow$ ResNet50": attacking WideResNet while generating adversarial samples from ResNet50; ``PSPNet $\rightarrow$ DeepLabv3": attacking PSPNet while obtaining adversarial samples with DeepLabv3; ``SSD $\rightarrow$ RFBNet": attacking SSD while acquiring adversarial samples from RFBNet.}} 
	\caption{Results of the ablation study for loss functions.} 
	\vspace{-0.1in}
	\Huge
	\label{tab:black-box-1-14}
	\resizebox{1.0\linewidth}{!}{
		\begin{tabular}{l|p{3.5cm}<{\centering}p{3.5cm}<{\centering}p{3.5cm}<{\centering}p{3.5cm}<{\centering}|p{3.5cm}<{\centering}p{3.5cm}<{\centering}p{3.5cm}<{\centering}p{3.5cm}<{\centering}|p{3.5cm}<{\centering}p{3.5cm}<{\centering}p{3.5cm}<{\centering}p{3.5cm}<{\centering}}
			\toprule[1pt]
			\multirow{2}{5cm}{WideResNet $\rightarrow$ ResNet50}&\multicolumn{4}{c|}{CIFAR10 (Accuracy \%)} &\multicolumn{4}{c|}{CIFAR100 (Accuracy \%)} &\multicolumn{4}{c}{Tiny-ImageNet (Accuracy \%)}  \\
			\cline{2-13}
			& {clean} & {PGD} & {DeepFool}& {C\&W} & {clean} & {PGD} & {DeepFool}& {C\&W} & {clean} & {PGD} & {DeepFool}& {C\&W}   \\
			\hline
			Ours w/o $\mathcal{L}_r$&87.3&72.8&86.7&87.1&61.8&48.9&60.9&61.6&62.7&42.6&58.8 &60.6  \\
			Ours w/o $\mathcal{L}_p$ &88.4 &78.6&87.1&87.5&63.5&52.8&62.4&64.7&63.1&44.2&60.4 &62.0  \\
			Ours w/o $\mathcal{L}_m$ &90.2&80.1&89.7&90.2&67.3&53.6&67.0&67.2&62.2&45.8&61.7 &62.5  \\
			Ours w/o $\mathcal{L}_{GAN_g}$& 87.0&73.4&81.6&82.0&62.1&50.4&61.6&62.3&60.6&40.3&57.2 &58.4  \\
			Ours w/o $\mathcal{L}_{F_{task}}$&92.6 &62.5&85.5&85.8&70.5&40.2&66.3&67.0&63.3&35.5&57.9 &58.7  \\
			Ours w/o $\mathcal{L}_{F_{rec}}$&91.8 &60.7&82.3&83.1&68.6&43.7&64.2&65.3&62.8&38.2&56.3 &57.9  \\
			Ours w/o $\mathcal{L}_{F_{align}}$& 90.3&70.6&83.0&84.4&70.5&47.2&62.2&63.0&62.7&41.1&55.3 &56.8  \\
			Ours w/o $\mathcal{L}_{F_{inter}}$&91.1 &73.8&85.2&86.7&69.4&49.1&63.9&64.7&61.8&42.5&57.0 &57.6  \\
			Ours w/o $\mathcal{L}_{F_{intra}}$& 90.8&71.3&86.1&87.2&70.8&47.9&65.1&66.0&62.5&44.1&59.2 &60.5  \\
			Full&90.7&\textbf{81.4}&\textbf{90.4}&\textbf{90.6}&68.7&\textbf{54.0}&\textbf{67.5}&\textbf{68.4}&62.9&\textbf{46.4} &\textbf{62.2} &\textbf{63.2}\\
			\hline
			PSPNet $\rightarrow$ DeepLabv3&\multicolumn{4}{c|}{Cityscapes (mIoU \%)} &\multicolumn{4}{c|}{VOC2012 (mIoU \%)} &\multicolumn{4}{c}{VOC07+12 (mAP \%)}  \\
			\cline{2-13}
			SSD $\rightarrow$ RFBNet& {clean} & {BIM} & {DeepFool}& {C\&W} & {clean} & {BIM} & {DeepFool}& {C\&W} & {clean} & {cls} & {loc} & {cls+loc}    \\
			\hline
			Ours w/o $\mathcal{L}_r$&61.5&50.8&52.4&56.1&66.4&54.3&56.2&61.6&56.2&58.3&55.6 &55.8  \\
			Ours w/o $\mathcal{L}_p$ &64.6 &53.2&57.3&59.5&69.1&58.2&61.4&63.7&57.8&59.5&56.2 &57.2  \\
			Ours w/o $\mathcal{L}_m$ & 67.2&57.8&61.3&62.4&70.1&62.4&64.5&67.6&59.6&60.1&57.3 &58.5  \\
			Ours w/o $\mathcal{L}_{GAN_g}$&60.2 &51.3&53.1&53.9&67.3&55.6&58.3&59.0&55.8&56.3&53.1 &53.7  \\
			Ours w/o $\mathcal{L}_{F_{task}}$& 64.5&47.7&51.2&53.6&72.6&48.2&59.1&61.8&63.5&53.6&55.0 &56.4  \\
			Ours w/o $\mathcal{L}_{F_{rec}}$&63.7&50.2&53.4&55.0&72.0&51.3&57.6&58.2&62.3&50.4&52.7 &53.4  \\
			Ours w/o $\mathcal{L}_{F_{align}}$&64.5 &51.4&55.7&57.2&71.6&57.8&59.3&60.1&60.3&55.2&51.8 &53.0  \\
			Ours w/o $\mathcal{L}_{F_{inter}}$&64.8 &52.5&56.8&58.3&70.8&59.2&60.5&62.6&59.7&57.6&52.4 &55.8  \\
			Ours w/o $\mathcal{L}_{F_{intra}}$&65.1 &54.7&58.2&60.8&70.5&61.0&61.2&64.3&60.4&58.2&54.3 & 57.0 \\
			Full&67.6&\textbf{59.5}&\textbf{62.0}&\textbf{64.7}&71.2 &\textbf{63.6} &\textbf{66.0}& \textbf{69.5} &60.5& \textbf{61.2}&\textbf{58.3} &\textbf{60.9} \\
			%%%%%%%%%%%%%%%%%%%%%%%%%%%%%%%%%%%%%%%
			\bottomrule[1pt]
			%\hline
	\end{tabular}}
\vspace{-0.1in}
\end{table*}

\subsection{Ablation Study}
%%\vspace{0.05in}
\noindent\textbf{Novel pixel-level constraints.} 
For each task, we conduct an ablation study to analyze the impact of each loss term and verify the superiority on pixel- and feature-level constraints.
We denote $\mathcal{L}_I$ as the results with only pixel-level constraints in our approach, $\mathcal{L}_{I+F(w/o \ c)}$ as the scores with our full constraints apart from $\mathcal{L}_{F_{class}}$. $\mathcal{L}_{T}$ represents the consequence with only traditional pixel-level constraints, and $\mathcal{L}_{T+F(w/o \ c)}$ refers to the results with traditional pixel- and feature-level constraints other than $\mathcal{L}_{F_{class}}$. The results are reported in Table \ref{tab:black-box}.

Compared with traditional pixel-level constraints, our novel pixel-level alignment strategy achieves much higher robustness on adversarial samples with a negligible degradation on clean samples as cost. Such superiority is prominent within the comparison between $\mathcal{L}_I$ and $\mathcal{L}_T$, $\mathcal{L}_{I+F(w/o \ c)}$ and $\mathcal{L}_{T+F(w/o \ c)}$. 
Such superiority demonstrates that adversarial samples can be better aligned with clean samples, when we match them with clean ones in the output space of the generator.

\vspace{0.05in}
\noindent\textbf{Integrated distribution alignment.} 
Further, compared to $\mathcal{L}_{I+F(w/o \ c)}$, our full setting exhibits stable improvement for effects on clean samples and adversarial samples (as shown in Table \ref{tab:black-box}), manifesting the impact of our proposed $\mathcal{L}_{F_{class}}$. %The positive effects of $\mathcal{L}_{F_{class}}$ are derived from its 
The positive effects of $\mathcal{L}_{F_{class}}$ prove the importance of aligning the overall distribution in the feature space of the target model.

\vspace{0.05in}
\noindent\textbf{Loss functions.}
There are several loss terms in Eq.~\ref{loss_over}, and we conduct ablation studies to verify their importance by deleting different loss terms from Eq.~\ref{loss_over} individually. And the corresponding results are recorded in Table~\ref{tab:black-box-1-14}.

\vspace{0.05in}
\noindent\textbf{Alternative pixel-level loss terms.}
As shown in Fig.~\ref{fig:pix}, traditional and our pixel-level training constraints have different reconstruction loss and adversarial loss terms, and we set ablation study to analyze the effectiveness if we using the mix strategies.
Thus, there are two settings.

\vspace{0.05in}
\noindent I) We have the reconstruction loss in the traditional strategy ($\Vert \widehat{x}^c-x^c \Vert$ and $\Vert \widehat{x}^a-x^c \Vert$), and the adversarial loss in our strategy (Eq.~\ref{gan1} and Eq.~\ref{fm1}). Such an ablation setting is called ``Full-abla-pixel-I".

\vspace{0.05in}
\noindent II) Use the reconstruction loss in our strategy ($\Vert \widehat{x}^c-x^c \Vert$ and $\Vert \widehat{x}^a-\widehat{x}^c \Vert$), and the adversarial loss in the traditional strategy. Especially, the adversarial loss can be written as
%%%\begin{equation}
%%%\begin{aligned}
%%%\mathcal{L}_{{GAN}_{d}} = &\mathbb{E}_{x^c \sim \mathcal{C}} ((\mathcal{D}(x^c)-1)^2) +\\
%%%& \mathbb{E}_{x^c \sim \mathcal{C}} ((\mathcal{D}(\widehat{x}^c)-0)^2)+ \\
%%%& \mathbb{E}_{x^a \sim \mathcal{A}} ((\mathcal{D}(\widehat{x}^a)-0)^2)\\ 
%%%\mathcal{L}_{{GAN}_{g}} = &\mathbb{E}_{x^c \sim \mathcal{A}}((\mathcal{D}(\widehat{x}^c)-1)^2)+\\
%%%&\mathbb{E}_{x^a \sim \mathcal{A}}((\mathcal{D}(\widehat{x}^a)-1)^2),
%%%\end{aligned}
%%%\label{gan2}
%%%\end{equation}
%%%\begin{equation}
%%%\begin{aligned}
%%%\mathcal{L}_{{m}} =&\mathbb{E}(\Vert \mathcal{F}(\widehat{x}^a) -\mathcal{F}(x^c) \Vert_1)+\\
%%%&\mathbb{E}(\Vert \mathcal{F}(\widehat{x}^c) -\mathcal{F}(x^c) \Vert_1),
%%%\end{aligned}
%%%\label{fm2}
%%%\end{equation}
\begin{equation}
\small
\begin{aligned}
\mathcal{L}_{{GAN}_{d}} = &\mathbb{E}_{x^c \sim \mathcal{C}} ((\mathcal{D}(x^c)-1)^2) +\mathbb{E}_{x^c \sim \mathcal{C}} ((\mathcal{D}(\widehat{x}^c)-0)^2)+ \\
& \mathbb{E}_{x^a \sim \mathcal{A}} ((\mathcal{D}(\widehat{x}^a)-0)^2)\\ 
\mathcal{L}_{{GAN}_{g}} = &\mathbb{E}_{x^c \sim \mathcal{A}}((\mathcal{D}(\widehat{x}^c)-1)^2)+\mathbb{E}_{x^a \sim \mathcal{A}}((\mathcal{D}(\widehat{x}^a)-1)^2),
\end{aligned}
\label{gan2}
\end{equation}
\begin{equation}
\small
\begin{aligned}
\mathcal{L}_{{m}} =&\mathbb{E}(\Vert \mathcal{F}(\widehat{x}^a) -\mathcal{F}(x^c) \Vert_1)+\mathbb{E}(\Vert \mathcal{F}(\widehat{x}^c) -\mathcal{F}(x^c) \Vert_1),
\end{aligned}
\label{fm2}
\end{equation}
and such an ablation setting is called ``Full-abla-pixel-II".
The corresponding results are shown in Table~\ref{tab:black-box-1-14-2}.

\begin{table*}[t]
	\centering
	%\caption{\xgxu{Results of the ablation study. ``WideResNet $\rightarrow$ ResNet50": attacking WideResNet while generating adversarial samples from ResNet50; ``PSPNet $\rightarrow$ DeepLabv3": attacking PSPNet while obtaining adversarial samples with DeepLabv3; ``SSD $\rightarrow$ RFBNet": attacking SSD while acquiring adversarial samples from RFBNet.}} 
	\caption{Results of the ablation study for ``alternative pixel-level loss terms", ``alternative feature-level loss terms", and ``the layer choice in the discriminator".} 
	\vspace{-0.1in}
	\Huge
	\label{tab:black-box-1-14-2}
	\resizebox{1.0\linewidth}{!}{
		\begin{tabular}{l|p{3.5cm}<{\centering}p{3.5cm}<{\centering}p{3.5cm}<{\centering}p{3.5cm}<{\centering}|p{3.5cm}<{\centering}p{3.5cm}<{\centering}p{3.5cm}<{\centering}p{3.5cm}<{\centering}|p{3.5cm}<{\centering}p{3.5cm}<{\centering}p{3.5cm}<{\centering}p{3.5cm}<{\centering}}
			\toprule[1pt]
			\multirow{2}{5cm}{WideResNet $\rightarrow$ ResNet50}&\multicolumn{4}{c|}{CIFAR10 (Accuracy \%)} &\multicolumn{4}{c|}{CIFAR100 (Accuracy \%)} &\multicolumn{4}{c}{Tiny-ImageNet (Accuracy \%)}  \\
			\cline{2-13}
			& {clean} & {PGD} & {DeepFool}& {C\&W} & {clean} & {PGD} & {DeepFool}& {C\&W} & {clean} & {PGD} & {DeepFool}& {C\&W}   \\
			\hline
			Full-abla-pixel-I& 89.7&77.3&87.3&88.0&67.6&50.7&64.4&65.2&63.4&45.1&59.4&60.3  \\
			Full-abla-pixel-II&89.4 &76.7&86.8&87.3&67.9&51.2&65.0&65.4&63.2&44.9&60.3&61.4   \\
			Full-abla-feature-I& 90.6&79.2&85.9&86.2&68.8&53.5&67.1&67.6&62.7&46.0&60.1&61.2   \\
			Full-abla-feature-II&90.5 &80.5&89.1&89.5&68.0&52.6&66.3&68.0&62.1&45.7&58.6& 59.7  \\
			deleting 1-st layer & 89.2&80.1&88.3&88.7&67.3&52.8&66.2&67.5&61.4&45.2&60.8 &62.1  \\
			deleting 2-nd layer &90.1 &79.6&87.5&88.0&66.8&50.3&64.8&64.2&62.1&44.5& 60.2&61.5  \\
			deleting 3-rd layer &88.3 &79.1&87.0&86.8&66.2&53.1&64.0&65.1&61.8&43.2& 58.7&60.3  \\
			deleting 4-th layer &87.8 &77.3&85.3&85.6&65.7&51.8&62.8&63.7&60.4&43.8& 58.0&60.6  \\
			deleting 5-th layer & 89.5&78.9&86.2&87.2&67.1&52.5&63.7&64.9&62.3&44.9&59.3 & 59.8 \\
			Full&90.7&\textbf{81.4}&\textbf{90.4}&\textbf{90.6}&68.7&\textbf{54.0}&\textbf{67.5}&\textbf{68.4}&62.9&\textbf{46.4} &\textbf{62.2} &\textbf{63.2}\\
			\hline
			PSPNet $\rightarrow$ DeepLabv3&\multicolumn{4}{c|}{Cityscapes (mIoU \%)} &\multicolumn{4}{c|}{VOC2012 (mIoU \%)} &\multicolumn{4}{c}{VOC07+12 (mAP \%)}  \\
			\cline{2-13}
			SSD $\rightarrow$ RFBNet& {clean} & {BIM} & {DeepFool}& {C\&W} & {clean} & {BIM} & {DeepFool}& {C\&W} & {clean} & {cls} & {loc} & {cls+loc}    \\
			\hline
			Full-abla-pixel-I&64.7 &55.4&57.0&60.5&71.4&58.3&63.2&67.6&59.2&58.7& 54.4&57.2  \\
			Full-abla-pixel-II&64.0 &55.1&56.8&58.7&70.8&57.1&62.8&66.5&59.5&59.0& 55.1& 56.4 \\
			Full-abla-feature-I&66.1 &58.2&59.4&61.6&72.0&61.2&64.1&68.3&60.7&60.6&57.5 & 60.1 \\
			Full-abla-feature-II&65.6 &57.9&58.5&59.2&72.2&60.5&63.7&67.8&60.4&59.3&56.8 & 59.5 \\
			deleting 1-st layer &67.0 &58.4&61.2&63.8&70.4&62.7&65.2&68.4&60.1&60.4& 57.2&60.3  \\
			deleting 2-nd layer &66.3 &57.5&60.8&63.2&70.8&61.8&64.3&67.5&60.8&58.8&55.6 &59.4  \\
			deleting 3-rd layer &67.1 &57.0&61.7&62.7&71.6&60.4&63.8&67.1&59.7&58.1&56.2 &58.8  \\
			deleting 4-th layer & 65.8&55.8&59.4&60.3&71.4&60.1&61.5&65.6&59.2&56.5& 54.7&56.2  \\
			deleting 5-th layer &66.7 &56.3&60.5&61.9&70.1&61.3&63.1&64.9&60.3&57.0&55.0 & 57.6 \\
			Full&67.6&\textbf{59.5}&\textbf{62.0}&\textbf{64.7}&71.2 &\textbf{63.6} &\textbf{66.0}& \textbf{69.5} &60.5& \textbf{61.2}&\textbf{58.3} &\textbf{60.9} \\
			%%%%%%%%%%%%%%%%%%%%%%%%%%%%%%%%%%%%%%%
			\bottomrule[1pt]
			%\hline
	\end{tabular}}
\vspace{-0.1in}
\end{table*}

\vspace{0.05in}
\noindent\textbf{Alternative feature-level loss terms.}
In the feature-level alignment, we can also utilize the similar strategy in the pixel-level alignment: using the features of $\widehat{z}^c$ to guide the formulation of $\widehat{z}^a$.
To demonstrate the drawback of such alternative strategy, we set two experiments.

\noindent I) we modify Eq.~\ref{feature-rec} to the following form, as
\begin{equation}
\begin{aligned}
&\widehat{z}^a=\mathcal{O}(\widehat{x}^a), \ \widehat{z}^c=\mathcal{O}(\widehat{x}^c), \ z^c=\mathcal{O}(x^c), \\ &\mathcal{L}_{{F}_{rec}}=\mathbb{E}(\Vert \widehat{z}^c-z^c \Vert_1) + \mathbb{E}(\Vert \widehat{z}^a-\widehat{z}^c \Vert_1),\\
\end{aligned}
\label{feature-rec2}
\end{equation}
and we keep other loss terms unchanged. Such an ablation setting is called ``Full-abla-feature-I".

\noindent II) We use the distribution of $\widehat{z}^c$ to align the distribution of $\widehat{z}^a$ and change Eq.~\ref{feature-align} to the following equation, as 
\begin{equation}
\begin{aligned}
\mathcal{L}_{{F}_{align}}&=\sum_{k=1:K} \mathbb{E}(\Vert \widehat{m}^{c(k)}-\widehat{m}^{a(k)}\Vert_1),\\
\end{aligned}
\label{feature-align2}
\end{equation}
where $\widehat{m}^{c(k)}$ denotes the clustering center of $\widehat{z}^{c(k)}$, and we preserve the other loss terms not modified, and called such an ablation setting as ``Full-abla-feature-II".
The corresponding results are shown in Table~\ref{tab:black-box-1-14-2}.

\vspace{0.05in}
\noindent\textbf{The layer choice in the discriminator.}
In Eq.~\ref{fm1}, we have employed the loss to measure the distances between clean and adversarial samples in terms of the discriminative feature maps which are extracted from the discriminator. 
We adopt the ``MultiscaleDiscriminator" following the setting of pix2pixHD \cite{wang2018high}. And there are 5 layers in the discriminator and we choose all these layers to compute $\mathcal{L}_m$ since the corresponding results are optimal.
To prove this, we deleting individual layer from computing $\mathcal{L}_m$ (denoted as ``deleting 1-st/2-nd/3-rd/4-th/5-th layer"), and we provide the experimental results in Table~\ref{tab:black-box-1-14-2} to demonstrate that the layer selected by us is optimal.

\begin{table*}[t]
	\centering
	\caption{Results of changing hyper-parameters $\lambda_1 \sim \lambda_4$ in Eq. \ref{loss_over}. The evaluation setting is the same as that of Table \ref{tab:black-box0} and \ref{tab:black-box1}.} 
	%\huge
	\vspace{-0.1in}
	\Huge
	\label{tab:hyper}
	\resizebox{1.0\linewidth}{!}{
		\begin{tabular}{l|p{3.5cm}<{\centering}p{3.5cm}<{\centering}p{3.5cm}<{\centering}p{3.5cm}<{\centering}|p{3.5cm}<{\centering}p{3.5cm}<{\centering}p{3.5cm}<{\centering}p{3.5cm}<{\centering}|p{3.5cm}<{\centering}p{3.5cm}<{\centering}p{3.5cm}<{\centering}p{3.5cm}<{\centering}}
			\hline
			&\multicolumn{4}{c|}{CIFAR10 (Accuracy \%)} &\multicolumn{4}{c}{CIFAR100 (Accuracy \%)} &\multicolumn{4}{c|}{Tiny-ImageNet (Accuracy \%)}\\
			\cline{2-13}
			& {clean} & {PGD} & {DeepFool}& {C\&W} & {clean} & {PGD} & {DeepFool}& {C\&W} & {clean} & {PGD} & {DeepFool}& {C\&W}  \\
			\hline
			$(\lambda_1 \sim \lambda_4)$ $\times$ 0.1 &86.6&78.2&87.7&88.2&66.4&51.2&65.3&64.2&60.3&43.8&59.4&60.1\\
			$(\lambda_1 \sim \lambda_4)$ $\times$ 10 &85.3&77.5&86.9&87.4&65.8&53.0&66.1&66.7&61.2&44.1&60.5&61.4\\
			original $(\lambda_1 \sim \lambda_4)$ &\textbf{90.7}&\textbf{81.4}&\textbf{90.4}&\textbf{90.6}&\textbf{68.7}&\textbf{54.0}&\textbf{67.5}&\textbf{68.4}&\textbf{62.9}&\textbf{46.4}&\textbf{62.2}&\textbf{63.2}\\
			\hline
			&\multicolumn{4}{c}{Cityscapes (mIoU \%)} &\multicolumn{4}{c|}{VOC2012 (mIoU \%)} &\multicolumn{4}{c}{VOC07+12 (mAP \%)} \\
			\cline{2-13}
			& {clean} & {BIM} & {DeepFool}& {C\&W}& {clean} & {BIM} & {DeepFool}& {C\&W}& {clean} & {cls} & {loc} & {cls+loc}  \\
			\hline
			$(\lambda_1 \sim \lambda_4)$ $\times$ 0.1  &63.5&56.3&58.4&60.4&69.3&61.4&62.6&67.8&56.6&57.7&54.1&57.2\\
			$(\lambda_1 \sim \lambda_4)$ $\times$ 10 &64.7&54.7&60.8&63.6&67.1&60.7&61.5&65.1&57.9&58.3&57.0&58.1\\
			original $(\lambda_1 \sim \lambda_4)$&\textbf{67.6}&\textbf{59.5}&\textbf{62.0}&\textbf{64.7}&\textbf{71.2}&\textbf{63.6}&\textbf{66.0}&\textbf{69.5}&\textbf{60.5}&\textbf{61.2}&\textbf{58.3}&\textbf{60.9}\\
			\hline
	\end{tabular}}
%	\vspace{-0.1in}
\end{table*}

\begin{table}[t]
	\centering
	\caption{Comparison among our approach, existing methods and ablation setting on the classification task, under the evaluation of model transfer setting. ``WideResNet $\Rightarrow$ ResNet50" means that the generator is trained with the target model of WideResNet, while we evaluate its defense effect for ResNet50.} 
	\vspace{-0.1in}
	\label{tab:diff-box1}
	\large
	\resizebox{1.0\linewidth}{!}{
		\begin{tabular}{l|ccc|ccc}
			%\hline
			\toprule[1pt]
			\multirow{2}{2.7cm}{WideResNet $\Rightarrow$ ResNet50}&\multicolumn{3}{c|}{CIFAR10 (Accuracy \%)} &\multicolumn{3}{c}{CIFAR100 (Accuracy \%)} \\
			\cline{2-7}
			& {PGD} & {DeepFool}& {C\&W} & {PGD} & {DeepFool}& {C\&W}  \\
			\hline
			No Defense &8.2&4.3&5.8& 17.3 &18.3&16.6   \\
			{$\mathcal{L}_I$} &44.5&84.6&85.7& 29.6 &63.4&64.2   \\
			{$\mathcal{L}_{I+F(w/o \ c)}$}&67.1&88.2&88.6&40.1 &64.1& 65.5  \\
			{$\mathcal{L}_T$}&18.1&84.0&84.7&14.9 &62.6&63.4   \\
			{$\mathcal{L}_{T+F(w/o \ c)}$}&59.9&87.0&87.2&36.6 &63.2&65.2   \\
			\hline
			{Defense \cite{samangouei2018defense}} &38.5&37.8&38.9 &21.5 &22.8&22.5   \\
			{SR \cite{mustafa2019image}} &44.8&45.2&45.3 &24.8& 25.5&25.1 \\
			{FPD \cite{li2020enhancing}}&44.2&44.0&44.4&30.5 &31.1&31.2  \\
			{APE \cite{shen2017ape}} &10.1&86.2&86.4&11.4 &64.8&65.4   \\
			{Denoise \cite{liao2018defense}} &73.0&87.7&87.1 &35.7 &63.7&64.2  \\
			{NRP \cite{naseer2020self}} &60.4&87.3&88.0& 36.2&63.5&64.6   \\
			Ours &\textbf{73.3}&\textbf{88.4}&\textbf{89.1}&\textbf{42.0} &\textbf{64.9}&\textbf{65.8}  \\
			\bottomrule[1pt]
			%\hline
	\end{tabular}}
%	\vspace{-0.1in}
\end{table}

\begin{table}[t]
	\centering
	\caption{Comparison among our approach, existing methods and ablation settings on the semantic segmentation task, under the evaluation of model transfer setting. ``PSPNet $\Rightarrow$ DeepLabv3" means the generator is trained with the target model of PSPNet, while we evaluate its defense effect for DeepLabv3.} 
	\vspace{-0.1in}
	\label{tab:diffbox2}
	\large
	\resizebox{1.0\linewidth}{!}{
		\begin{tabular}{l|ccc|ccc}
			%\hline
			\toprule[1pt]
			\multirow{2}{2.7cm}{PSPNet $\Rightarrow$ DeepLabv3}&\multicolumn{3}{c|}{Cityscapes (mIoU \%)} &\multicolumn{3}{c}{VOC2012 (mIoU \%)}  \\
			\cline{2-7}
			& {BIM} & {DeepFool}& {C\&W} & {BIM} & {DeepFool}& {C\&W} \\
			\hline
			No Defense &3.8&32.3&13.5&11.8&49.0&20.2 \\
			$\mathcal{L}_I$ &48.5&49.9&55.0&54.9&59.8&68.0 \\
			$\mathcal{L}_{I+F(w/o \ c)}$&57.8&60.3&62.2&59.9&63.2&68.2  \\
			$\mathcal{L}_T$&43.0&48.9&53.7&33.2&49.9&53.7  \\
			$\mathcal{L}_{T+F(w/o \ c)}$&53.8&58.6&60.0&55.1&60.3&65.8  \\
			\hline
			Defense \cite{samangouei2018defense} &20.1&19.7&20.7&21.6&21.2&23.1  \\
			SR \cite{mustafa2019image} &41.6&40.5&42.5&52.1&56.9&66.8 \\
			FPD \cite{li2020enhancing}&51.5&51.7&53.6&56.7&57.3&60.5  \\
			APE \cite{shen2017ape}&28.5&26.8&40.0&25.7&46.9&44.5 \\
			Denoise \cite{liao2018defense}&52.7&51.1&63.4&60.1&53.7&67.2  \\
			NRP \cite{naseer2020self}  &53.4&47.8&63.0&53.1&51.2&67.4  \\
			Ours &\textbf{59.0}&\textbf{61.7}&\textbf{64.3}&\textbf{60.5}&\textbf{64.6}&\textbf{68.9} \\
			\bottomrule[1pt]
	\end{tabular}}
%	\vspace{-0.1in}
\end{table}

\subsection{Hyper-parameters Analysis}
The hyper-parameters of our method include the loss weights in Eq. \ref{loss_over}.
If values of all loss weights are set to tenfold or tenth, the results are shown in Table \ref{tab:hyper} (the evaluation setting is the same as that of Table \ref{tab:black-box0} and \ref{tab:black-box1}).
This table demonstrate that results with original hyper-parameters are the highest. And accuracy in the classification task, mIoU in the segmentation task, mAP in the detection task are altered no more than 6.1\%, 8.0\%, 7.2\%.
Thus, our model's effect is not sensitive to hyper-parameters, and our chosen parameters are reasonable.

\begin{table*}[t]
	\centering
	%\caption{\xgxu{Quantitative comparison on the classification tasks with targeted attack. 
			%	``WideResNet $\rightarrow$ ResNet50": attacking WideResNet while generating adversarial samples from ResNet50}.} 
	\caption{Quantitative comparison on the classification tasks with the targeted attack.}
	\vspace{-0.1in}
	\Huge
	\label{tab:black-box0-target}
	\resizebox{1.0\linewidth}{!}{
		\begin{tabular}{l|p{3.5cm}<{\centering}p{3.5cm}<{\centering}p{3.5cm}<{\centering}p{3.5cm}<{\centering}|p{3.5cm}<{\centering}p{3.5cm}<{\centering}p{3.5cm}<{\centering}p{3.5cm}<{\centering}|p{3.5cm}<{\centering}p{3.5cm}<{\centering}p{3.5cm}<{\centering}p{3.5cm}<{\centering}}
			\toprule[1pt]
			\multirow{2}{5cm}{WideResNet $\rightarrow$ ResNet50}&\multicolumn{4}{c|}{CIFAR10 (Accuracy \%)} &\multicolumn{4}{c|}{CIFAR100 (Accuracy \%)} &\multicolumn{4}{c}{Tiny-ImageNet (Accuracy \%)}  \\
			\cline{2-13}
			& {clean} & {PGD} & {DeepFool}& {C\&W} & {clean} & {PGD} & {DeepFool}& {C\&W} & {clean} & {PGD} & {DeepFool}& {C\&W}   \\
			\hline
			%%No Defense & 95.1&22.2&25.6&27.4&78.1&32.6&34.5&37.2&\textbf{64.5}&42.2&44.4 &45.1  \\
			%%No Defense (finetune) & \textbf{95.6} &22.8&26.3&28.5&\textbf{79.5}&33.2&34.9&37.8&63.9&42.3&44.8 &45.7  \\
			No Defense & 95.1&1.6&3.8&5.1&78.1&7.1&7.7&9.1&\textbf{64.5}&17.3&17.7 &19.6  \\
			No Defense (finetune) & \textbf{95.6} &3.1&4.8&6.5&\textbf{79.5}&7.6&8.0&9.4&63.9&20.6&21.1 &21.7  \\
			\hline
			TRADES \cite{zhang2019theoretically} &87.3  &76.3&85.4&85.7&62.8&51.2&60.4&60.1&58.5&42.7&55.6&55.9  \\
			TRADES (finetune) \cite{zhang2019theoretically}  & 85.4&75.0&83.1&83.8&78.1&51.0&52.7&43.6&43.3&40.1&41.7 &42.0  \\
			Free-adv \cite{shafahi2019adversarial} & 77.1 &70.2&72.6&74.3&49.2&43.8&45.8&46.8&55.5&43.2& 53.5&53.8  \\
			Free-adv (finetune) \cite{shafahi2019adversarial} & 88.5 &77.6&86.3&87.6&63.7&52.6&61.5&61.9&42.2&40.5&46.0 &46.3  \\
			\hline
			Defense \cite{samangouei2018defense} &39.9&36.2&36.5&37.0&31.1&28.4&28.8&30.3&20.4&16.3&17.4 &17.8  \\
			SR \cite{mustafa2019image}&48.0&45.8&47.7&47.5&33.6&31.8&32.1&32.5&31.1&29.4&29.7 &30.0  \\
			FPD \cite{li2020enhancing}&48.5&45.7&47.2&47.8&52.5&40.3&41.0&41.4&39.7&32.7&37.1 &38.1  \\
			APE \cite{shen2017ape} &90.2&40.5&88.4&89.2&73.2&35.5&63.6&66.2&62.4&29.9&60.3 &60.7  \\
			Denoise \cite{liao2018defense}&89.8&78.4&88.6&89.1&67.2&51.7&65.2&65.7&59.8&44.6&56.8 &57.2  \\
			NRP \cite{naseer2020self} &91.8&76.7&87.9&88.4&70.3&50.2&65.4&66.0&59.1&43.0&56.6 &57.5  \\
			Ours &90.7&\textbf{80.1}&\textbf{88.8}&\textbf{89.7}&68.7&\textbf{53.1}&\textbf{66.3}&\textbf{66.7}&62.9&\textbf{44.8}&\textbf{61.2} &\textbf{62.4}  \\
			\bottomrule[1pt]
	\end{tabular}}
	\vspace{-0.1in}
\end{table*}

\begin{table*}[t]
	\centering
	%\caption{\xgxu{Quantitative comparison on the classification tasks with AT that trains with feature-level constraints. 
			%	``WideResNet $\rightarrow$ ResNet50": attacking WideResNet while generating adversarial samples from ResNet50.} }
	\caption{Quantitative comparison on the classification tasks with AT that trains with feature-level constraints. And ``+Ours" means the setting that replaces the corresponding original feature-level constraints to our feature-level constraints.}
	\vspace{-0.1in}
	\Huge
	\label{tab:black-box0-transfer}
	\resizebox{1.0\linewidth}{!}{
		\begin{tabular}{l|p{3.5cm}<{\centering}p{3.5cm}<{\centering}p{3.5cm}<{\centering}p{3.5cm}<{\centering}|p{3.5cm}<{\centering}p{3.5cm}<{\centering}p{3.5cm}<{\centering}p{3.5cm}<{\centering}|p{3.5cm}<{\centering}p{3.5cm}<{\centering}p{3.5cm}<{\centering}p{3.5cm}<{\centering}}
			\toprule[1pt]
			\multirow{2}{5cm}{WideResNet $\rightarrow$ ResNet50}&\multicolumn{4}{c|}{CIFAR10 (Accuracy \%)} &\multicolumn{4}{c|}{CIFAR100 (Accuracy \%)} &\multicolumn{4}{c}{Tiny-ImageNet (Accuracy \%)}  \\
			\cline{2-13}
			& {clean} & {PGD} & {DeepFool}& {C\&W} & {clean} & {PGD} & {DeepFool}& {C\&W} & {clean} & {PGD} & {DeepFool}& {C\&W}   \\
			\hline
			No Defense & 95.1&2.1&5.3&6.4&78.1&7.5&9.3&10.4&\textbf{64.5}&19.2&20.4 &21.2  \\
			No Defense (finetune) & \textbf{95.6} &3.8 &7.5&8.7 &\textbf{79.5}&8.0& 9.7&10.9&63.9&22.1&23.2&23.6 \\
			\hline
			PCL~\cite{mustafa2019adversarial}&88.7 &80.3&87.2&87.8&65.4&56.2&64.7&65.0&60.2&45.8&59.1 &60.3  \\
			PCL~\cite{mustafa2019adversarial}+Ours&88.4 &81.0(+0.7)&88.3(+1.1)&88.7(+0.9)&64.6&57.1(+0.9)&65.4(+0.7)&65.8(+0.8)&60.9&46.2(+0.4)&60.4(+1.3) & 61.9(+1.6) \\
			\hline
			CADA~\cite{hou2020class}&83.8 &74.6&83.6&84.3&60.8&52.6&60.1&60.9&55.7&42.0& 57.0& 57.8 \\
			CADA~\cite{hou2020class}+Ours&84.2 &75.1(+0.5)&85.2(+1.6)&85.7(+1.4)&61.3&53.3(+0.7)&60.9(+0.8)&61.7(+0.8)&56.2&43.1(+1.1) & 57.6(+0.6)& 58.4(+0.6)\\
			\hline
			ATDA~\cite{song2019improving}&87.5 &78.2&86.7&87.0&63.5&55.8&62.7&63.2&58.0.&43.7&58.4 &59.1  \\
			ATDA~\cite{song2019improving}+Ours&87.9 &79.5(+1.3)&87.5(+0.8)&87.9(+0.9)&64.2&56.4(+0.6)&63.8(+1.1)&64.5(+1.3)&58.7&44.5(+0.8)&59.5(+1.1) & 60.5(+1.4) \\
			\hline
			Ours &90.7&\textbf{81.4}&\textbf{90.4}&\textbf{90.6}&68.7&\textbf{54.0}&\textbf{67.5}&\textbf{68.4}&62.9&\textbf{46.4} &\textbf{62.2} &\textbf{63.2}\\
			\bottomrule[1pt]
	\end{tabular}}
	\vspace{-0.1in}
\end{table*}

\subsection{Experiments under Model Transfer Evaluation}
\label{sec:eva_transfer}
For defense via input transformation with deep generative models, we use a target model $\mathcal{O}$ for feature-level training. 
It has been verified that the trained generator $\mathcal{G}$ yields excellent defense quality for $\mathcal{O}$. 
On the other hand, $\mathcal{G}$ can be deployed as a plug-and-play module to safeguard different target models $\mathcal{O}'$ that were not employed during training. 
Evaluation with the model transfer setting \cite{liao2018defense} demonstrates this property.
In the classification task, we defend ResNet50 against black-box attack while the generator $\mathcal{G}$ is trained with the target model of WideResNet; as for semantic segmentation, $\mathcal{G}$ is trained with PSPNet when utilized for the protection of DeepLabv3.
The results are summarized in Tables \ref{tab:diff-box1} and \ref{tab:diffbox2}. These outcomes provide empirical conclusion that the trained generator $\mathcal{G}$ is applicable to the safeguard of the target model that is not adopted during training, and can outperform ablation settings and most existing methods.

\subsection{The Evaluation for Targeted Attack}
\label{sec:targeted}
For image classification, another major attack is the targeted attack. And we provide the evaluation under the targeted attack in this section. 
Compared with untargeted attack, the targeted adversarial samples aim to fool the classifier by outputting specific labels, and the target labels can be specified  by the adversary.
%%We chose the target label for each testing sample via random sampling to achieve the targeted attack. 
Following the setting in~\cite{croce2020reliable}, the targeted ones are 9 target classes attaining the 9 highest scores at the original point (except the correct one).
And all approaches are evaluated with the same setting.
% of the target label.
The results on CIFAR10, CIFAR100 and Tiny-ImageNet are shown in Table~\ref{tab:black-box0-target}.
And we can still see the advantages of our framework compared with baselines, further demonstrating the effectiveness of our approach.

\subsection{Applying Our Feature-level Constraints in AT}
Some existing adversarial training approaches have also considered the feature level alignment.
However, as discussed in Sec.~\ref{relate}, these methods~\cite{mustafa2019adversarial,mustafa2020deeply,hou2020class,song2019improving} can not achieve the overall distribution alignment and the alignment for paired samples (the adversarial sample and the corresponding clean sample) simultaneously.
To demonstrate the superiority of our feature-level constraints that can achieve integrated distribution alignment, we replace the feature-level loss in the frameworks of~\cite{mustafa2019adversarial,mustafa2020deeply,hou2020class,song2019improving} with ours, and observe the change of the performance.
To unify the training and evaluation setting, we retrain their models on different datasets with our chosen attack settings during training and evaluation.
As shown in Table~\ref{tab:black-box0-transfer}, after change the feature-level loss into our feature-level constraints, the performances of the corresponding methods are increased, while still lower than our framework's performance (``Ours").

\subsection{What If The Attacker Knows About Defense}
In this section, we analyze the situation where the attacker knows the existence of defense as described in \cite{naseer2020self}. In this situation, the attacker accesses the training data and mechanism, and trains a local defense similar to our trained $\mathcal{G}$, and adopts BPDA \cite{athalye2018obfuscated} to bypass the defense.
To simulate this attack, we adopt \cite{ronneberger2015u} as the structure of $\mathcal{G}$ and train it with our training mechanism defined in Eq. \eqref{loss_over}.

Besides, we apply PGD (translation-invariant attack), BIM, and ``cls+loc" along with the BPDA to implement attacks.
The accuracy of our framework on CIFAR10, CIFAR100 and Tiny-ImageNet (with WideResNet) under this attack setting are 74.5\%, 48.8\%, 40.3\%, and the results of NRP are 72.4\%, 45.1\%, 38.2\% respectively; the mIoU of our apporach on Cityscapes and VOC2012 (with PSPNet) are 53.6\%, 56.2\%, and the results with NRP are 50.7\%, 53.2\% respectively; the mAP of our method on VOC07+12 (with SSD) is 53.9\% while the mAP with NRP is 51.1\%. Obviously, BPDA cannot circumvent our defense and our defense outperforms NRP under this challenging setting.

\section{Conclusion}
In this paper, we have proposed a novel training scheme for DGNs that align the distribution of adversarial samples to clean samples for a given target model.
Effectiveness of our strategy stems from the pixel- and feature-level constraints. As a general approach, our framework is suitable for various tasks, including image classification, semantic segmentation, and object detection. Extensive experiments reveal the effect of our novel constraints and illustrate the advantage of our method compared with existing state-of-the-art defense strategies.

\ifCLASSOPTIONcompsoc
%\section*{Acknowledgments}
%\else
%\section*{Acknowledgment}
\fi


\ifCLASSOPTIONcaptionsoff
\newpage
\fi



%\bibliographystyle{IEEEtran}
%\begin{thebibliography}{1}


%\end{thebibliography}
\bibliographystyle{IEEEtran}
\bibliography{egbib}


\begin{IEEEbiography}
	[{\includegraphics[height=1.25in,clip,keepaspectratio]{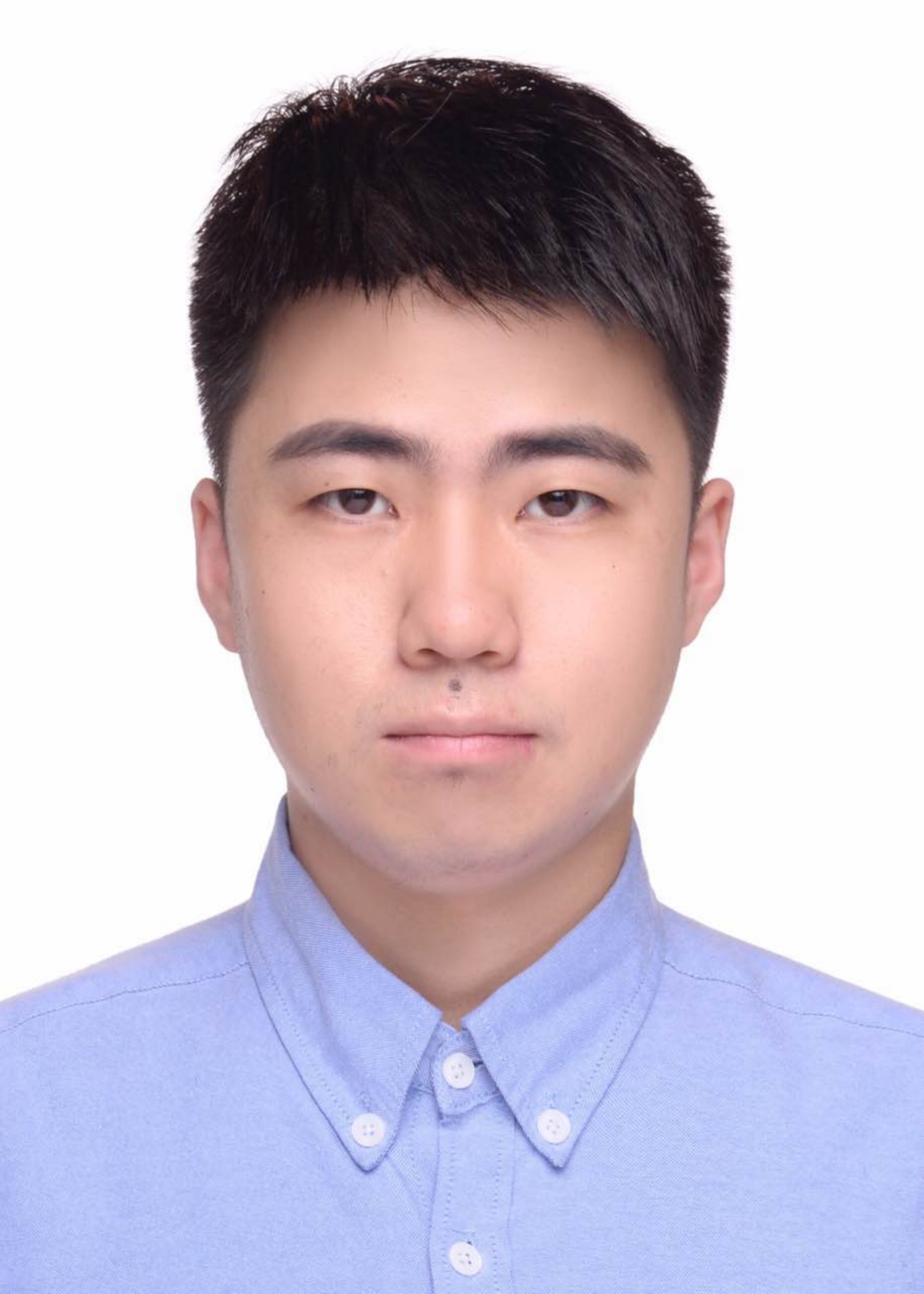}}]
	{Xiaogang Xu}
	is currently a fourth-year PhD student in the Chinese University of Hong Kong. He received his bachelor degree from Zhejiang University.
	He obtained the Hong Kong PhD Fellowship in 2018. He serves as a reviewer for CVPR, ICCV, ECCV, AAAI, ICLR, NIPS, IJCV. His research interest includes deep learning, generative adversarial networks, adversarial attack and defense, etc.
\end{IEEEbiography}

\begin{IEEEbiography}
	[{\includegraphics[height=1.25in,clip,keepaspectratio]{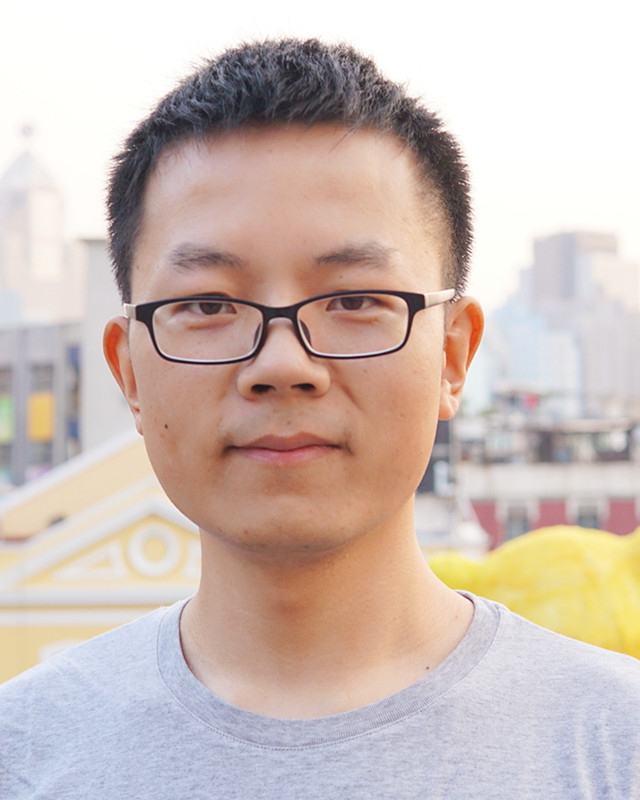}}]{Hengshuang Zhao} is currently an Assistant Professor in the Department of Computer Science at The University of Hong Kong. He received the PhD degree in Computer Science and Engineering from The Chinese University of Hong Kong. He worked as a postdoctoral researcher at the University of Oxford and Massachusetts Institute of Technology. He and his team won several champions in competitive academic challenges like ImageNet Scene Parsing, LSUN Semantic Segmentation, WAD Drivable Area Segmentation, Embodied AI Social Navigation, etc. His general research interests cover the broad area of computer vision and machine learning, with special emphasis on high-level scene recognition and pixel-level scene understanding. He is a member of the IEEE.
\end{IEEEbiography}

\begin{IEEEbiography}
	[{\includegraphics[height=1.25in,clip,keepaspectratio]{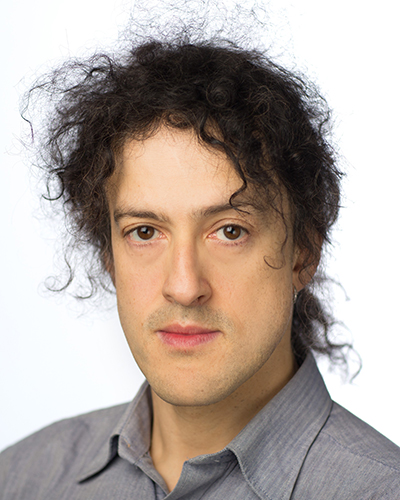}}]
	{Philip Torr} received the PhD degree from the University of Oxford. After working for another three years at Oxford as a research fellow, he worked for six years in Microsoft Research, first in Redmond, then in Cambridge, founding the vision side of the Machine Learning and Perception Group. He then became a Professor in Computer Vision and Machine Learning at Oxford Brookes University. He is now a professor at Oxford University. He is a BMVA Distinguished Fellow, Ellis Fellow, Royal Academy of Engineering Fellow, Royal Society Fellow, and Turing AI World-Leading Researcher Fellow.
\end{IEEEbiography}

%\vspace{-.2in}
\begin{IEEEbiography}
	[{\includegraphics[height=1.25in,clip,keepaspectratio]{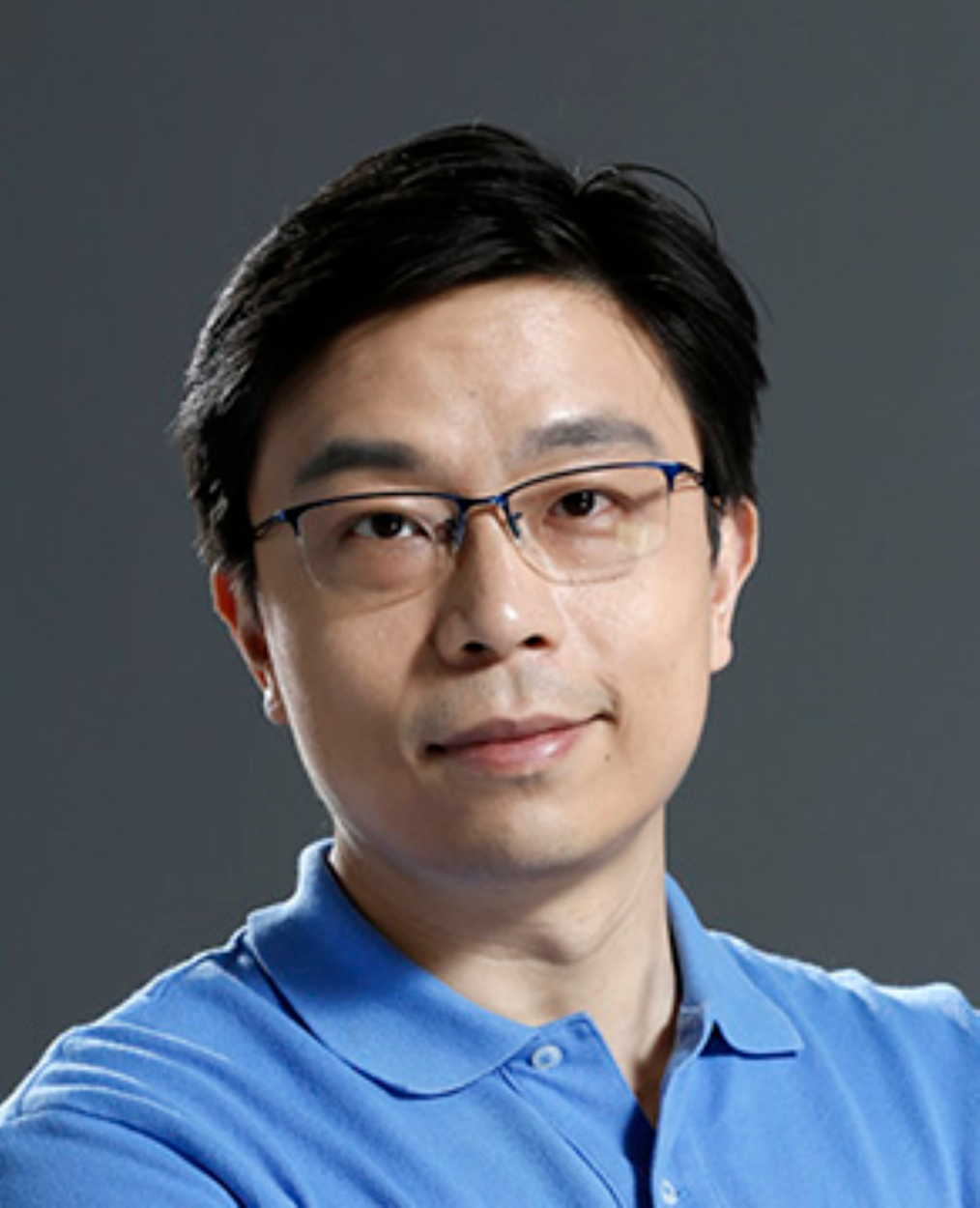}}]
	{Jiaya Jia}
	received the PhD degree in Computer Science from Hong Kong University of Science and Technology in 2004 and is currently a full professor in Department of Computer Science and Engineering at the Chinese University of Hong Kong (CUHK). He was a visiting scholar at Microsoft Research Asia from March 2004 to August 2005 and conducted collaborative research at Adobe Systems in 2007. He is an Associate Editor-in-Chief of IEEE Transactions on Pattern Analysis and Machine Intelligence (TPAMI) and  is also in the editorial board of International Journal of Computer Vision (IJCV). He continuously served as area chairs for ICCV, CVPR, AAAI, ECCV, and several other conferences for organization. He was on program committees of major conferences in graphics and computational imaging, including ICCP, SIGGRAPH, and SIGGRAPH Asia. He received the Young Researcher Award 2008 and Research Excellence Award 2009 from CUHK. He is a Fellow of the IEEE.
\end{IEEEbiography}

\end{document}